\documentclass[a4paper,fleqn]{cas-sc}
\usepackage[numbers,sort&compress]{natbib}
\usepackage{amsmath}
\usepackage{amssymb}
\usepackage{booktabs}
\usepackage{multirow}
\usepackage[section]{placeins}  

\makeatletter
\let\PI@origFloatBarrier\FloatBarrier
\renewcommand\FloatBarrier{%
  \PI@origFloatBarrier
  \ifx\@dbldeferlist\@empty\else\clearpage\fi}
\makeatother

\begin{document}
\let\WriteBookmarks\relax
\renewcommand{\dbltopfraction}{0.9}       
\renewcommand{\topfraction}{0.9}
\renewcommand{\bottomfraction}{0.5}
\renewcommand{\textfraction}{0.07}        
\renewcommand{\floatpagefraction}{0.7}    
\renewcommand{\dblfloatpagefraction}{0.6}

\shortcites{Bradbury2018,Rahman2024,Hao2023,Vaswani2017,Li2021,Park2026b,Zhao2023,%
Kissas2022,Liu2026b,Kushwaha2024,Park2026a,Hossain2025,Lu2021,Koric2024,Sahin2024}

\shorttitle{Disentangling attention in deep operator learning}
\shortauthors{A.A. Koric et~al.}

\title[mode = title]{Disentangling Attention in Deep Operator Learning:
  A Controlled Study of Data-Driven and Physics-Informed Architectures}

\author[2]{Amar Alem Koric}
\ead{koric@stanford.edu}
\credit{Methodology, Formal analysis, Investigation, Writing -- original draft}

\author[1]{Qibang Liu}
\ead{qibang@illinois.edu}
\credit{Conceptualization, Methodology, Writing -- review \& editing}

\author[1,3]{Seid Koric}
\cormark[1]
\ead{koric@illinois.edu}
\credit{Methodology, Supervision, Resources, Writing -- original draft, Writing -- review \& editing}

\affiliation[1]{organization={National Center for Supercomputing Applications,
    University of Illinois at Urbana-Champaign},
    addressline={1205 W. Clark St.},
    city={Urbana},
    state={IL},
    postcode={61801},
    country={USA}}

\affiliation[2]{organization={Department of Mechanical Engineering,
    Stanford University},
    city={Stanford},
    state={CA},
    postcode={94305},
    country={USA}}

\affiliation[3]{organization={The Grainger College of Engineering, Mechanical
    Science and Engineering, University of Illinois at Urbana-Champaign},
    addressline={1206 W. Green St.},
    city={Urbana},
    state={IL},
    postcode={61801},
    country={USA}}

\cortext[cor1]{Corresponding author}

\begin{abstract}
Deep neural operators learn mappings between input functions and complete PDE solution fields, enabling forward evaluations of new problem instances orders of magnitude faster than conventional numerical solvers. Attention mechanisms have recently been introduced into neural operators, but most studies change several architectural components at once, making it difficult to identify what actually improves accuracy. This work presents a controlled and systematic study of five deep operator network (DeepONet) variants with distinct attention mechanisms, trained under both data-driven and physics-informed regimes, to isolate the effects of cross-attention, self-attention, tokenization, and attention depth. We evaluate them on a source-driven transient one-dimensional nonlinear diffusion--reaction equation, a transient one-dimensional viscous Burgers equation with variable initial conditions, and a two-dimensional Poisson heat-conduction problem with heterogeneous source fields. Per-sensor tokenization with cross-attention reduces the mean relative $L_2$ error of the classical DeepONet in all benchmark--training combinations by factors of 2.4--28.0, while the best attention configurations reach 3.5--32.3. Branch self-attention paired only with dot-product fusion is inconsistent, degrading the one-dimensional problems while helping the more complex two-dimensional source field; added on top of cross-attention it improves all six cases, though by less than cross-attention fusion alone. Global pre-mixing provides no consistent benefit. Increasing cross-attention depth further improves accuracy, but with diminishing returns and a substantially higher cost under physics-informed training. Overall, query-dependent cross-attention is the most reliable mechanism, whereas branch self-attention is most useful for large, spatially complex functional inputs.
\end{abstract}


\maketitle

\section{Introduction and previous work}
\label{sec:intro}

High-fidelity numerical simulation is central to heat transfer, fluid
mechanics, solid mechanics, and many other engineering disciplines. In most
practical workflows, however, the computational challenge is not a single
forward solution, but the repeated solution of the same governing partial
differential equations (PDEs) as initial or boundary conditions, source
distributions, geometries, material properties, or operating parameters change.
Such repeated evaluations underpin sensitivity analysis, uncertainty
quantification, inverse design, digital-twin updating, and online monitoring
and control. When the governing system is transient, nonlinear, multiphysics,
or multiscale, the accumulated cost of these forward solves can dominate the
overall workflow, even on modern high-performance computing platforms, making
it challenging and often impractical.

Neural operators address this bottleneck by learning mappings between function
spaces rather than a single finite-dimensional input--output relation. Once
trained, they can predict solution fields for previously unseen inputs drawn
from the training distribution without rerunning the full numerical solver. Two
widely used formulations are the Fourier Neural Operator (FNO) and the Deep
Operator Network (DeepONet). FNO performs global spectral mixing efficiently on
rectangular domains with regular grids \citep{Li2021}, while Geo-FNO extends the
approach to irregular geometries through a learned deformation between the
physical domain and a uniform latent grid \citep{Li2023}. DeepONet, motivated by
the universal approximation theorem for nonlinear operators
\citep{ChenChen1995,Lu2021}, uses a branch network to encode samples of an input
function and a trunk network to encode a spatial or spatiotemporal query
coordinate. A terminal inner product combines the two latent representations.
This separation is attractive because it cleanly distinguishes conditioning data
from query location and allows the learned output field to be evaluated at
arbitrary coordinates.

Physics-informed DeepONet (PI-DeepONet) retains this operator-learning structure
while augmenting the training objective with the governing PDE and its initial
and boundary conditions \citep{Wang2021}. Because the predicted field is
differentiable with respect to the trunk coordinates, the required derivatives
can be evaluated by automatic differentiation and inserted directly into the PDE
residual. The resulting model can reduce the amount of labeled simulation data
required for training and, in some settings, can be trained primarily or
entirely from physical constraints.

DeepONet and its variants have consequently been applied across a broad range of
scientific and engineering problems. Representative examples include nonlinear
solid mechanics and inelastic material behavior
\citep{Koric2024,Park2026b,ZhongMeidani2024}, fracture mechanics
\citep{Goswami2022}, aerodynamics \citep{Zhao2023}, acoustics \citep{Xu2023},
seismology \citep{Haghighat2024}, heat transfer and thermal design
\citep{KoricAbueidda2023,Sahin2024}, coupled multiphysics and materials
processing \citep{Kushwaha2024,Park2026a}, and digital-twin-oriented prediction
and monitoring \citep{Kobayashi2024,Hossain2025}. These studies demonstrate the
practical value of learning solution operators when many related PDE instances
must be evaluated.

The standard DeepONet fusion is computationally efficient and provides a clear
separation between the input function and the output query, but its fixed
dot-product form may limit how adaptively these two representations interact.
For a given input function, the branch network produces a single latent
coefficient vector, while the trunk network produces a query-dependent latent
basis vector; their dot product then determines the prediction. Although the
query coordinate changes the trunk representation, it does not directly select
or reweight individual samples or localized features of the encoded input field.
At finite model capacity, this fixed bilinear interaction may be inefficient
when the influence of an input feature varies sharply with output location,
geometry, boundary-condition context, or long-range coupling. This practical
limitation does not contradict DeepONet's universal approximation properties;
rather, it motivates more adaptive mechanisms for branch--trunk interaction.

Attention provides a natural mechanism for such adaptive interaction. As a central component of the Transformer architecture introduced by
\cite{Vaswani2017}, attention has since become ubiquitous in modern deep learning, most prominently in large language models, and is increasingly being adopted for scientific machine learning. In cross-attention, a representation of the query
coordinate forms data-dependent weights over encoded conditioning tokens,
allowing the model to emphasize the portions of the input field that are most
relevant to that query. Self-attention serves a different purpose: it allows
field tokens to exchange information and represent nonlocal dependencies before
input--query fusion. These two mechanisms should therefore not be treated as
interchangeable. Their effects may also depend strongly on how input functions
are tokenized and how spatial, temporal, or geometric position is represented.

Attention-based operator learning has developed along several complementary
lines. The Galerkin Transformer adapted self-attention to operator learning and
interpreted simplified attention through a projection perspective
\citep{Cao2021}. Learning Operators with Coupled Attention (LOCA) used attention
weights that depend explicitly on output query locations, providing an early
example of query-adaptive operator evaluation \citep{Kissas2022}. The General
Neural Operator Transformer (GNOT) introduced heterogeneous normalized attention
and geometric gating to handle multiple input functions, irregular meshes, and
multiscale structure within a unified framework \citep{Hao2023}. Together, these
approaches show how attention can represent long-range interactions and
input-dependent coupling beyond a fixed terminal inner product.

Other architectures have changed the objects on which attention operates or the
way positional and geometric information enters the model. The Position-induced
Transformer (PiT) constructs attention from spatial relations between sampling
positions rather than from field values \citep{ChenWu2024}. The Codomain
Attention Neural Operator (CoDA-NO) tokenizes physical variables along the
codomain or channel dimension and extends positional encoding, self-attention,
and normalization to function spaces \citep{Rahman2024}. Geometry-informed
Neural Operator Transformer (GINOT) encodes unordered, nonuniform,
variable-size boundary point clouds and integrates geometric context with
solution queries through attention \citep{Liu2026b}. For sequential functional
inputs, Sequential Neural Operator Transformer (S-NOT) combines recurrent
encoding with self- and cross-attention to model interactions between loading
histories and spatial queries \citep{Liu2026a,Park2026b}. These developments
illustrate the flexibility of attention, but they also introduce distinct
tokenization, positional, geometric, and temporal modeling choices.

Attention has also entered physics-informed operator learning.
\citet{Foster2025} incorporated physics-informed training into GNOT and compared
the resulting model with a data-driven counterpart. PINTO uses iterative
cross-attention-based kernel-integral units to condition domain-query
representations on initial and boundary functions and is trained without labeled
simulations \citep{BoyaSubramani2025}. These studies establish the feasibility of
physics-informed transformer neural operators, including query-conditioned
cross-attention, but they do not isolate the contribution of attention from the
other architectural and training changes introduced at the same time.

This lack of isolation remains an important methodological gap. Most published
comparisons apply cross-attention alongside self-attention, as well as specific
tokenization, positional or geometric encoding, network depth and width, decoder
structure, or the training objective. Consequently, an observed accuracy gain
cannot be attributed unambiguously to cross-attentive branch--trunk fusion,
self-attention among input tokens, a particular tokenization scheme, increased
capacity, or another accompanying design choice. The attribution problem is even
more consequential in physics-informed operator learning, where changes in
attention representation alter automatic-differentiation paths and interact with
the relative weighting of PDE-residual, initial-condition, and
boundary-condition losses.

The present work addresses this attribution problem through a controlled,
component-level study of attention in deep operator learning. To the best of our
knowledge, prior work has not systematically isolated these mechanisms within
matched DeepONet-style architectures under both data-driven and physics-informed
training. Starting from classical DeepONet and PI-DeepONet baselines, we
gradually introduce the architectural changes: cross-attention for branch--trunk
fusion, self-attention within encoded field tokens, per-input versus global
tokenization, and cross-attention depth. Model capacity, optimizer settings,
sampling budgets, training schedules, and evaluation procedures are kept as
consistent as possible so that performance differences can be attributed to the
tested component rather than to a redesign of another bundled attention-based
operator architecture.

The study is organized around four practical questions: whether cross-attentive
fusion improves on the classical dot product; whether self-attention provides
additional value once query-dependent fusion is present; how per-input and
global tokenization alter accuracy and cost; and how cross-attention depth influences operator learning. We evaluate
predictive error together with training and inference cost. The
evaluation spans three engineering operators with distinct functional inputs and
governing physics: (1) transient diffusion-reaction driven by a variable spatial
source function; (2) transient viscous Burgers transport driven by a variable initial
condition, in which nonlinear self-advection steepens smooth data into sharp
interior fronts; and (3) thermal conduction governed by the Poisson
equation with a spatially complex two-dimensional heat-source distribution. The
third case extends the multidimensional source-field setting of
\citet{KoricAbueidda2023}, but the purpose here is different: it provides a
controlled test bed for determining which attention and coordinate-representation
choices are responsible for any improvement. Together, the three cases span
source-driven and initial-condition-driven operators, transient diffusion,
nonlinear advective transport, and strongly heterogeneous multi--dimensional spatial forcing. These
characteristics are representative of many real-world thermal and fluid
applications, including nonuniform or nonlinear heat generation in nuclear fuel
rods, localized heat sources in advanced chip manufacturing, and nonlinear wave
steepening and front propagation in compressible and reacting flows. Thus, rather
than proposing another externally perceived monolithic transformer operator, the novelty of this work
is a matched computational-experimental framework that disentangles the effects
of attention mechanisms, tokenization, and cross-attention depth and provides practical
evidence for when their added computational cost is justified.


\section{Formulations}
\label{sec:design}

\subsection{Operator learning background}
\label{sec:background}

Let $\Omega \subset \mathbb{R}^{d}$ denote the computational domain and 
$\partial\Omega$ its boundary, where $d$ is the spatial dimension. Let
$\mathcal{A}=\{a:\Omega\rightarrow\mathbb{R}^{c_{a}}\}$ denote the input functional
space and $\mathcal{U}=\{u:\Omega\rightarrow\mathbb{R}^{c_{u}}\}$ denote the
solution functional space, where $c_{a}$ and $c_{u}$ are the numbers of components of
the input and solution fields, respectively.

For each input function $a\in\mathcal{A}$, such as a boundary or initial condition,
distribution of source or material property, the corresponding solution $u\in\mathcal{U}$ satisfies
\begin{equation}
\mathcal{N}(u;a)=0
\quad \text{in } \Omega,
\qquad
u=g
\quad \text{on } \partial\Omega,
\label{eq:governing_pde}
\end{equation}
where $\mathcal{N}$ is the governing partial differential equation and $g$ is the
prescribed boundary condition.

The PDE solution operator $\mathcal{G}$ maps the input function space
$\mathcal{A}$ to the solution function space $\mathcal{U}$:
\begin{equation}
\mathcal{G}:\mathcal{A}\rightarrow\mathcal{U},
\qquad
a(\cdot)\mapsto u(\cdot)=\mathcal{G}(a).
\label{eq:solution_operator}
\end{equation}

Operator learning seeks to approximate this mapping by using a neural operator
$\mathcal{G}_{\theta}$, parameterized by trainable parameters $\theta$, such
that
\begin{equation}
u_{\theta}(x)
=
\mathcal{G}_{\theta}(a)(x),
\qquad
x\in\Omega.
\label{eq:learned_operator}
\end{equation}

\subsection{Two design axes}
\label{sec:design:axes}

Attention enters an operator network at more than one place, and published
architectures typically change several of them together. To attribute an accuracy
change to a specific mechanism, we organise the comparison along two axes that can be
varied independently: how the sampled input function is encoded, and how that encoding
is combined with the query representation.

The encoder axis, also referred to as the branch, governs the representation of the input function. In the classical formulation, the (m) sensor readings are concatenated into a single vector and passed through a multilayer perceptron, producing a single latent vector $(b(a)\in\mathbb{R}^{p})$. In transformer-based architectures, the input sequence is instead embedded into small numerical units, or tokens, and positional information is added to identify the sensor location. The branch therefore returns a sequence of tokens $(B(a)\in\mathbb{R}^{m\times p})$. A further level of representation applies self-attention across these tokens, allowing each sensor token to incorporate information from all other sensors. This enables dependencies between distant regions of the input field to be captured before fusion with the decoder representation.

Self-attention is restricted to the branch sensor tokens. Applying it across trunk queries would couple the output locations, causing each prediction to depend on the other points evaluated with it. This would violate the pointwise DeepONet representation and make predictions dependent on the chosen output discretization. Pointwise independence is also important for physics-informed training, where collocation points are resampled at each step. The trunk therefore remains pointwise throughout.

The fusion axis governs how the branch and trunk representations are combined to produce a prediction. In the classical DeepONet, the branch compresses all $m$ sensor measurements into a single vector $b(a)\in\mathbb{R}^{p}$, while the trunk represents the evaluation coordinate as $\tau(x)\in\mathbb{R}^{p}$. The prediction is then obtained through the inner product $\langle b(a),\tau(x)\rangle$. Although $\tau(x)$ depends on the query location $x$, the way the two representations are combined does not change: the model always takes a fixed inner product between the same $p$-dimensional latent components of $b(a)$ and $\tau(x)$. More importantly, the model cannot explicitly determine which sensors are most relevant to a particular query location.

Cross-attention instead preserves the branch representation as a sequence of sensor tokens and uses the trunk representation as a query. For each evaluation coordinate $x$, the model computes a different set of attention weights over the sensor tokens. It can therefore emphasize sensors that are more informative for that location while reducing the influence of less relevant sensors. In this way, cross-attention provides a location-dependent fusion mechanism, whereas the classical inner product combines a fixed global summary of the input function with the trunk representation.

The five variants examined here are listed in~\ref{sec:design:variants}. They are designed to provide focused comparisons of the main architectural choices under study, including tokenization, branch self-attention, global branch information, and the fusion rule. Where possible, comparisons change one main mechanism at a time, while related choices such as positional encoding are kept consistent within each benchmark and are not varied independently in this study.

Finally, the goal of this study is not to compare the attention-based variants with every modern neural operator architecture. Instead, we use a common DeepONet framework, under both data-driven and physics-informed training, to determine which attention mechanisms are responsible for the observed improvements. This controlled setting allows us to study the role of attention itself without mixing its effects with other major architectural differences.

\subsection{The five architectural variants}
\label{sec:design:variants}

\begin{figure*}[t]
\centering
\includegraphics[width=0.75\textwidth]{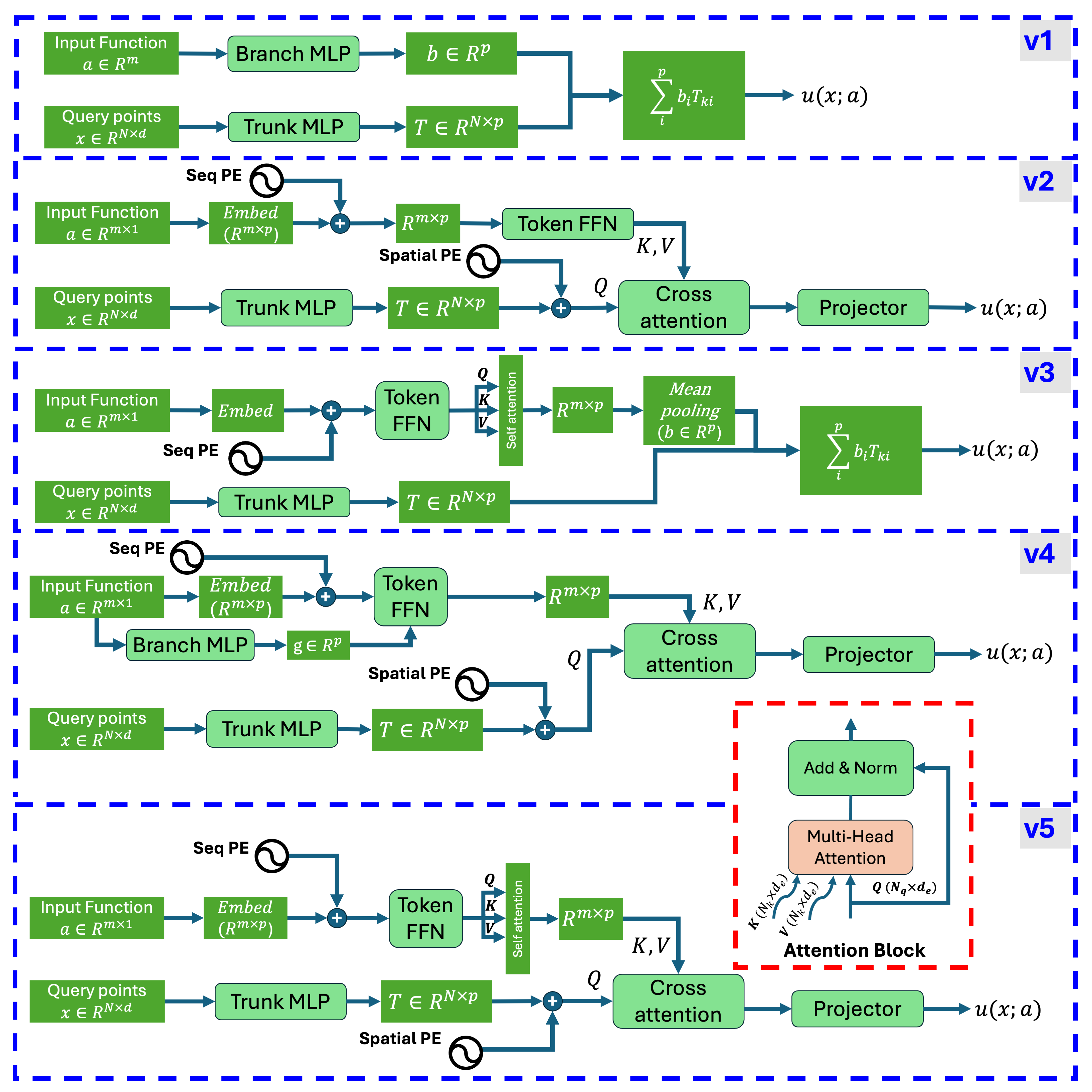}
\caption{The five variants, one per blue rectangle. The red panel defines the attention block used wherever attention appears above: multi-head attention followed by a residual addition and a layer normalization. Cross-attention takes the query from the trunk and the keys and values from the branch tokens, with the residual carrying the projected query $Q$. Self-attention forms all three from the branch tokens and takes the residual from the token stream before those projections.}
\label{fig:architectures}
\end{figure*}

Figure~\ref{fig:architectures} compares the five variants considered in this study. V1 is the standard DeepONet architecture and serves as the baseline. V2 represents the branch input as sensor tokens and replaces dot-product fusion with cross-attention, testing the combined effect of tokenization and cross-attentive fusion. V3 applies self-attention to the branch tokens before pooling them into a global representation and retaining the standard dot-product fusion, thereby isolating the effect of self-attention in the branch encoder. V4 extends V2 with a branch MLP that injects global information about the input function into each sensor token before branch-trunk cross-attention. Finally, V5 combines branch self-attention with branch-trunk cross-attention to form the complete attention-based architecture.

Several components recur across these variants. The embedding layer (Embed) maps each scalar sensor value to the latent dimension ($p$), representing an input function as ($m$) tokens rather than a single global vector. The token feedforward network (Token FFN) then applies a two-layer nonlinear transformation with GELU activation to each token independently, enriching its representation before attention. Following cross-attention, the (Projector) maps the fused representation to the scalar prediction through a residual addition and LayerNorm, followed by a two-layer MLP with GELU activation. The physics-informed Poisson models are an exception, as they omit LayerNorm from the readout and retain only the residual addition. Positional encodings (PE) are introduced at the locations indicated in Figure~\ref{fig:architectures}; their form depends on the input function and training paradigm, as described in Section~\ref{sec:design:pe}. The (Cross-attention) block also serves as the repeated unit in the depth study of Section~\ref{sec:exp:depth}, where (N) stacked blocks successively refine the query representation. All models outside that study use a single cross-attention block.

All models are implemented in JAX \citep{Bradbury2018}. Its automatic differentiation capability provides the derivatives of the predicted solution with respect to the query coordinates required to evaluate the physics-informed residuals.

\subsection{The attention block}
\label{sec:design:attn}

Both mechanisms examined in this study are built from the same operation and differ only in
where the query, key and value sequences are drawn from. Following \citet{Vaswani2017},
attention is evaluated with $h$ heads of width $d_{k}=p/h$,
\begin{equation}
\mathrm{Attn}(Q,K,V)
=
\mathrm{softmax}\!\left(\frac{QK^{\top}}{\sqrt{d_{k}}}\right)V ,
\label{eq:attention}
\end{equation}
where the projections $W_{Q},W_{K},W_{V}\in\mathbb{R}^{p\times p}$ that form $Q$, $K$ and $V$
are learned separately in each block.

Let $B(a)\in\mathbb{R}^{m\times p}$ denote the branch tokens and $\tau(x)$ the trunk
output at query coordinate $x$, to which the positional encoding $\gamma(x)$ introduced in
Section~\ref{sec:design:pe} is added. In cross-attentive fusion the query is formed from
that trunk output and the keys and values from the branch tokens,
\begin{equation}
Q = W_{Q}\big(\tau(x)+\gamma(x)\big),
\qquad
K = W_{K}B(a),
\qquad
V = W_{V}B(a),
\label{eq:qkv}
\end{equation}
and the residual is carried by the projected query $Q$. In branch self-attention the three
sequences are instead formed from the branch tokens alone,
\begin{equation}
Q = W_{Q}B(a),
\qquad
K = W_{K}B(a),
\qquad
V = W_{V}B(a),
\label{eq:qkv-sa}
\end{equation}
and the residual is taken from the token stream $B(a)$ before those projections. In both
cases the output of Equation~\eqref{eq:attention} is combined with its residual and passed
through a layer normalization. The two then diverge: self-attention returns an updated token
sequence, which continues into the fusion stage, whereas cross-attention returns the fused
representation itself, from which the prediction follows through the two-layer feedforward
read-out, as outlined in Figure~\ref{fig:architectures}.

The query dependence of cross-attentive fusion follows from Equations~\eqref{eq:attention}
and~\eqref{eq:qkv}: because the query is formed from the trunk output at the evaluation
coordinate, together with its positional encoding, the attention weights
$\mathrm{softmax}(QK^{\top}/\sqrt{d_{k}})$ vary with that coordinate. Consequently, different query locations can emphasize different parts of the encoded input field, a mechanism also used in query-conditioned operator architectures such as those of \citet{Kissas2022} and \citet{Hao2023}. Branch self-attention carries no such dependence, since Equation~\eqref{eq:qkv-sa} involves no query coordinate; it acts on the encoded input function before fusion.

The attention block defined by Equations~\eqref{eq:attention}--\eqref{eq:qkv-sa} is identical
in form, latent width, head count, scaling and residual structure across all three benchmarks
and both training paradigms as referenced in ~\ref{sec:design:protocol}. The read-out is likewise shared, with a single exception: the
physics-informed Poisson models omit its layer normalization and retain only the residual
addition, as noted in Section~\ref{sec:design:variants}.
\FloatBarrier

\subsection{Positional encoding of sensor tokens}
\label{sec:design:pe}

Once the branch is tokenized, attention treats the sensor readings as an unordered
collection: reordering the tokens leaves the result unchanged. The spatial arrangement of
the sensors therefore has to be supplied separately, by adding to each token a fixed
vector that encodes the location it came from. This information is known to be decisive
in attention-based operator learning, to the point that it can be used as the sole basis
for the attention weights \citep{ChenWu2024}. The encoding is applied to the branch tokens and to the
query in the three cross-attention variants V2, V4 and V5. In V3 it is applied only to the branch tokens, before self-attention, because
dot-product fusion gives the query no token representation for an encoding to act on.
The baseline V1 uses no positional encoding at all.

The scheme is matched to the input function and to the training paradigm rather than
held fixed across benchmarks. For the transient diffusion-reaction operator we use the sinusoidal encoding of
\citet{Vaswani2017} evaluated at the sensor and query coordinates themselves. It varies
smoothly with the coordinate, so it can also be used in the physics-informed models. For the transient Burgers operator, whose domain is periodic in space, the spatial
encoding uses integer harmonics $\sin(2\pi nx)$ and $\cos(2\pi nx)$ for $n=1,\dots,4$,
so that $x=0$ and $x=1$, which are the same physical point, receive identical encodings.
The time coordinate retains the sinusoidal form.

The Poisson benchmark needs a different treatment. Its input is a rough Gaussian random
field read at $1024$ sensors, so the encoding must clearly distinguish sensors that lie
close together. In the data-driven models we therefore evaluate the sinusoidal encoding
at the position of each sensor in the grid ordering, counted as an integer, rather than
at its physical coordinate. Because the sensor grid is fixed and identical for every
sample the two carry the same information, and the integer version separates
neighbouring sensors far more strongly.

This option is not open to the physics-informed models. Their queries are collocation
points that may lie anywhere in the domain, and the prediction must be differentiable
with respect to $(x,y)$ so that the residual in Equation~\eqref{eq:governing_pde} can be
formed by automatic differentiation. An encoding indexed by an integer cannot be
differentiated in this way. Those models therefore combine the two: the branch tokens
keep the integer-based encoding, while the query uses a sinusoidal encoding whose
frequencies are chosen so that the slowest completes one full oscillation across the
domain and the fastest completes $\mathrm{MAX\_CYCLES}$. This retains differentiability
while still separating sensors at the spacing of the grid.

The positional encodings were introduced to support effective learning on each benchmark,
but their type and placement were selected on a benchmark-specific basis and were not varied
systematically. The present results therefore do not isolate the effect of positional encoding.
A dedicated study comparing encoding types and their placement within the branch and trunk networks
would be required in a future work.

\subsection{Fair-comparison protocol}
\label{sec:design:protocol}

Within a given benchmark and training paradigm, the five variants share every setting
that is not the mechanism under test. They use the same latent width $p$,
number of attention heads wherever attention is present, trunk architecture and
activation, optimizer and learning-rate schedule, number of
iterations, sampling budget per step, and held-out inputs, so that all
comparisons are paired. These values are collected in Table~\ref{tbl:protocol}.

To ensure a controlled comparison, we keep the latent dimension $p$, which defines the representation used for branch-trunk fusion, fixed across all five architectural variants within each PDE benchmark. Artificially matching the total parameter count would require changing the width or depth of individual variants, introducing an additional architectural difference and confounding the comparison. The corresponding data-driven and physics-informed models are also architecturally matched and have identical parameter counts for the transient diffusion--reaction and transient Burgers problems, with only minor differences for the Poisson attention variants, as reported in Table~\ref{tbl:params}. The remaining variation in parameter count across V1--V5 arises naturally from the mechanisms under study: cross-attention introduces additional projections, while self-attention and token-processing layers further change the number of trainable parameters. We therefore hold the latent dimension fixed and explicitly report the resulting parameter counts rather than artificially equalizing them.

Every model in the study is trained with grouped sampling. Each step draws a set of input
functions and a set of output (collocation and boundary) points for each of them. The branch is evaluated once per
input function in the set,  and its encoded representation is reused across all its collocation and boundary points rather than evaluating the branch once per point, such as in flat/random batching. This shortens the training time and lowers the
memory required \citep{Mandl2025}, which is particularly useful for the attention variants.

\begin{table}[width=.95\linewidth,cols=4,pos=h]
\caption{Shared training and architecture settings. All five variants within a
given benchmark and paradigm use identical values.}
\label{tbl:protocol}
\small
\begin{tabular*}{\tblwidth}{@{\extracolsep{\fill}}lrrr@{}}
\toprule
Setting & Diff.-react. & Burgers & Poisson \\
\midrule
Latent dimension $p$      & 128 & 128 & 128 \\
Attention heads           & 4   & 4  & 4 \\
Trunk hidden layers       & $4\times128$ & $4\times128$ & $4\times128$ \\
Branch sensors $m$        & 100 & 101 & 1024 \\
Activation                & tanh & tanh & tanh \\
Training iterations       & 120{,}000 & 120{,}000 & 80{,}000 \\
Input functions per step  & 100 & 100 & 128 (DD) / 64 (PI) \\
Batch size (points/step)  & 10{,}000 & 10{,}000 & 4096 (DD) / 2048 (PI) \\
Optimizer                 & Adam & Adam & Adam \\
\bottomrule
\end{tabular*}
\end{table}

\begin{table*}[width=\textwidth,cols=5,pos=t]
\caption{Trainable parameter counts.}
\label{tbl:params}
\small
\begin{tabular*}{\tblwidth}{@{\extracolsep{\fill}}lrrrr@{}}
\toprule
 & Tr. diff.-react. & Tr. Burgers & \multicolumn{2}{c}{Poisson} \\
\cmidrule(lr){4-5}
Variant & DD $=$ PI & DD $=$ PI & DD & PI \\
\midrule
V1 dot product   & 112{,}384 & 112{,}512 & 230{,}656 & 230{,}656 \\
V2 tokens $+$ CA & 182{,}145 & 182{,}145 & 182{,}145 & 181{,}889 \\
V3 SA $+$ dot    & 165{,}504 & 165{,}504 & 165{,}504 & 165{,}504 \\
V4 global $+$ CA & 244{,}737 & 244{,}865 & 363{,}009 & 362{,}753 \\
V5 SA $+$ CA     & 231{,}553 & 231{,}553 & 231{,}553 & 231{,}297 \\
\bottomrule
\end{tabular*}
\end{table*}

\FloatBarrier

\section{Numerical experiments}
\label{sec:experiments}

\subsection{Evaluation protocol and overall standings}
\label{sec:exp:protocol}

The five variants are evaluated on three benchmark operators, selected to span different governing equations and input-function types, with each variant trained in both data-driven and physics-informed settings, for a total of 30 experiments. Poisson is elliptic and steady, with a two-dimensional
source read at $1024$ sensors; diffusion-reaction is parabolic and transient, driven by a
one-dimensional source fixed in time; Burgers is transient and advection dominated, and
takes an initial condition rather than a forcing term.
The data-driven models minimise the
mean squared error against the reference solution,
\begin{equation}
\mathcal{L}_{\mathrm{data}}(\theta)
=
\frac{1}{N_{s}}\sum_{i=1}^{N_{s}}
\frac{1}{P}\sum_{j=1}^{P}
\Big(\mathcal{G}_{\theta}(a_{i})(x_{ij}) - u_{i}(x_{ij})\Big)^{2},
\label{eq:mse}
\end{equation}
over the $N_{s}$ input functions and $P$ points per function drawn at each step. The
physics-informed models never see the reference solution during training. Following
\citet{Wang2021}, they minimise instead the sum of a residual term formed from
Equation~\eqref{eq:governing_pde} at interior collocation points and a term penalizing
departures from the initial and boundary conditions, weighted equally. The residual is
specific to each benchmark and is given with the corresponding problem setup.

All models are assessed by the relative $L_2$ error of the predicted field against the
reference, computed separately for each held-out input function,

\begin{equation}
L_{2}
=
\frac{
    \left\| \boldsymbol{u}^{\mathrm{true}}
    - \boldsymbol{u}^{\mathrm{pred}} \right\|_{2}
}{
    \left\| \boldsymbol{u}^{\mathrm{true}} \right\|_{2}
},
\label{eq:rell2}
\end{equation}
with both norms taken over the full evaluation grid. Training uses standardized inputs and
solutions, but Equation~\eqref{eq:rell2} is evaluated after the standardization is undone,
so the reported errors are in physical units.
Each of 30 experiments was independently repeated three times using different random seeds. For each run, the mean and standard deviation of the relative $L_2$ error were computed across the held-out test samples, and Table~\ref{tbl:main}  reports the averages of these run-wise means and standard deviations over the three runs in units of $10^{-2}$. Per-benchmark results follow in Sections~\ref{sec:exp:dr} to
\ref{sec:exp:poisson}.

\begin{table*}[width=\textwidth,cols=7,pos=t]
\caption{Mean relative $L_2$ error on held-out inputs $\pm$ standard deviation across
those inputs, in units of $10^{-2}$. The smaller number beneath each entry is the
improvement factor over the V1 dot-product baseline, i.e.\ the ratio of the V1 mean to
that variant's mean.
Best mean in each column in bold. ``Tr.'' denotes transient, i.e.\ the time-dependent
diffusion--reaction and Burgers benchmarks; the Poisson benchmark is steady.}
\label{tbl:main}
\small
\begin{tabular*}{\tblwidth}{@{\extracolsep{\fill}}lrrrrrr@{}}
\toprule
Variant & \multicolumn{6}{c}{Mean relative $L_2$ error \ ($\times 10^{-2}$)} \\
\cmidrule(lr){2-7}
 & \multicolumn{3}{c}{Data-driven} & \multicolumn{3}{c}{Physics-informed} \\
\cmidrule(lr){2-4}\cmidrule(lr){5-7}
 & Tr. diff.-react. & Tr. Burgers & Poisson & Tr. diff.-react. & Tr. Burgers & Poisson \\
\midrule
V1 dot product      & $1.38 \pm 0.727$ & $4.06 \pm 2.17$ & $29.5 \pm 17.6$ & $0.764 \pm 0.426$ & $12.4 \pm 7.10$ & $33.4 \pm 24.1$ \\
\addlinespace[5pt]
V2 tokens $+$ CA    & \begin{tabular}[t]{@{}r@{}}$0.346 \pm 0.160$\\[-1pt]{\scriptsize $4.0\times$}\end{tabular} & \begin{tabular}[t]{@{}r@{}}$0.753 \pm 0.508$\\[-1pt]{\scriptsize $5.4\times$}\end{tabular} & \begin{tabular}[t]{@{}r@{}}$1.05 \pm 0.613$\\[-1pt]{\scriptsize $28.0\times$}\end{tabular} & \begin{tabular}[t]{@{}r@{}}$0.312 \pm 0.156$\\[-1pt]{\scriptsize $2.4\times$}\end{tabular} & \begin{tabular}[t]{@{}r@{}}$1.58 \pm 1.06$\\[-1pt]{\scriptsize $7.9\times$}\end{tabular} & \begin{tabular}[t]{@{}r@{}}$6.15 \pm 2.77$\\[-1pt]{\scriptsize $5.4\times$}\end{tabular} \\
\addlinespace[5pt]
V3 SA $+$ dot       & \begin{tabular}[t]{@{}r@{}}$3.98 \pm 2.28$\\[-1pt]{\scriptsize $0.35\times$}\end{tabular} & \begin{tabular}[t]{@{}r@{}}$4.59 \pm 2.31$\\[-1pt]{\scriptsize $0.88\times$}\end{tabular} & \begin{tabular}[t]{@{}r@{}}$3.81 \pm 2.20$\\[-1pt]{\scriptsize $7.7\times$}\end{tabular} & \begin{tabular}[t]{@{}r@{}}$1.81 \pm 1.07$\\[-1pt]{\scriptsize $0.42\times$}\end{tabular} & \begin{tabular}[t]{@{}r@{}}$19.2 \pm 8.36$\\[-1pt]{\scriptsize $0.64\times$}\end{tabular} & \begin{tabular}[t]{@{}r@{}}$3.96 \pm 2.39$\\[-1pt]{\scriptsize $8.4\times$}\end{tabular} \\
\addlinespace[5pt]
V4 global $+$ CA    & \begin{tabular}[t]{@{}r@{}}$0.399 \pm 0.170$\\[-1pt]{\scriptsize $3.5\times$}\end{tabular} & \begin{tabular}[t]{@{}r@{}}$0.536 \pm 0.427$\\[-1pt]{\scriptsize $7.6\times$}\end{tabular} & \begin{tabular}[t]{@{}r@{}}$1.43 \pm 0.956$\\[-1pt]{\scriptsize $20.6\times$}\end{tabular} & \begin{tabular}[t]{@{}r@{}}$\mathbf{0.221 \pm 0.105}$\\[-1pt]{\scriptsize $3.5\times$}\end{tabular} & \begin{tabular}[t]{@{}r@{}}$\mathbf{0.704 \pm 0.679}$\\[-1pt]{\scriptsize $17.6\times$}\end{tabular} & \begin{tabular}[t]{@{}r@{}}$7.04 \pm 4.57$\\[-1pt]{\scriptsize $4.7\times$}\end{tabular} \\
\addlinespace[5pt]
V5 SA $+$ CA        & \begin{tabular}[t]{@{}r@{}}$\mathbf{0.319 \pm 0.147}$\\[-1pt]{\scriptsize $4.3\times$}\end{tabular} & \begin{tabular}[t]{@{}r@{}}$\mathbf{0.527 \pm 0.395}$\\[-1pt]{\scriptsize $7.7\times$}\end{tabular} & \begin{tabular}[t]{@{}r@{}}$\mathbf{0.913 \pm 0.522}$\\[-1pt]{\scriptsize $32.3\times$}\end{tabular} & \begin{tabular}[t]{@{}r@{}}$0.276 \pm 0.143$\\[-1pt]{\scriptsize $2.8\times$}\end{tabular} & \begin{tabular}[t]{@{}r@{}}$1.03 \pm 0.770$\\[-1pt]{\scriptsize $12.1\times$}\end{tabular} & \begin{tabular}[t]{@{}r@{}}$\mathbf{2.88 \pm 1.13}$\\[-1pt]{\scriptsize $11.6\times$}\end{tabular} \\
\bottomrule
\end{tabular*}
\end{table*}

Table~\ref{tbl:main} identifies the same broad pattern in
all six benchmark--training combinations. V5 (self-attention $+$ cross-attention)
gives the lowest errors for all three data-driven problems and for physics-informed
Poisson, whereas V4 (global branch $+$ cross-attention) performs best for
physics-informed transient diffusion--reaction and transient Burgers. More generally,
every architecture containing cross-attentive branch--trunk fusion, namely V2
(tokens $+$ cross-attention), V4 (global branch $+$ cross-attention), and V5
(self-attention $+$ cross-attention), improves upon V1 (the dot-product baseline) in
all six cases. In contrast, V3 (self-attention $+$ dot-product fusion) is less accurate
than V1 (the dot-product baseline) for both transient one-dimensional problems but is
substantially more accurate for the two-dimensional Poisson problem.

\FloatBarrier

\subsection{Transient diffusion-reaction with a variable source}
\label{sec:exp:dr}

\subsubsection{Problem setup}
\label{sec:exp:dr:setup}

The operator maps a spatially varying source term to the solution of a nonlinear
diffusion-reaction equation,
\begin{equation}
\frac{\partial u}{\partial t}
= \kappa \frac{\partial^{2} u}{\partial x^{2}} + k\,u^{2} + f(x),
\qquad (x,t)\in[0,1]^{2},
\label{eq:dr}
\end{equation}
with $\kappa = k = 0.01$ and homogeneous initial and boundary conditions. The input
function $a$ is the source $f$, drawn from a Gaussian process with a squared-exponential
kernel of length scale $0.2$ and sampled at $m=100$ equally spaced sensors. Reference
solutions are obtained on a $100\times100$ grid in $(x,t)$ with an implicit
finite-difference scheme \citep{Iserles2009}. We generate $5000$ source functions for
training and hold out $100$ further functions, drawn from the same distribution but from
a separate random stream, for evaluation. All five variants are trained and tested on the
same functions, so every comparison reported below is paired.

The data-driven models are fitted by minimizing the mean squared error against the
reference solution at output points sampled from each training function. The
physics-informed models never see the reference solution during training. Following
\citet{Wang2021}, their loss combines a residual term, which evaluates
Equation~\eqref{eq:dr} at interior collocation points using derivatives obtained by
automatic differentiation through the trunk coordinates, with a term that penalises
departures from the prescribed initial and boundary conditions. Each training step draws
$100$ source input functions and, for each, $100$ boundary points and $100$
collocation points, giving $10^{4}$ points per loss term per step. The two terms are
weighted equally. Reference solutions are used only to compute the errors reported in
Section~\ref{sec:exp:dr:pi}.

\subsubsection{Data-driven results}
\label{sec:exp:dr:dd}

Under supervised training, V1 (the dot-product baseline) gives a mean relative $L_2$
error of $1.38\times10^{-2}$ over the 100 held-out source functions. The three
cross-attention variants reduce this error to $3.46\times10^{-3}$ for V2
(tokens $+$ cross-attention), $3.99\times10^{-3}$ for V4
(global branch $+$ cross-attention), and $3.19\times10^{-3}$ for V5
(self-attention $+$ cross-attention). V5 (self-attention $+$ cross-attention) therefore
provides the lowest mean error, while V3
(self-attention $+$ dot-product fusion) is the only variant that performs worse than V1
(the dot-product baseline), with a mean error of $3.98\times10^{-2}$.

\begin{figure*}[tp]
\centering
\includegraphics[width=0.75\textwidth]{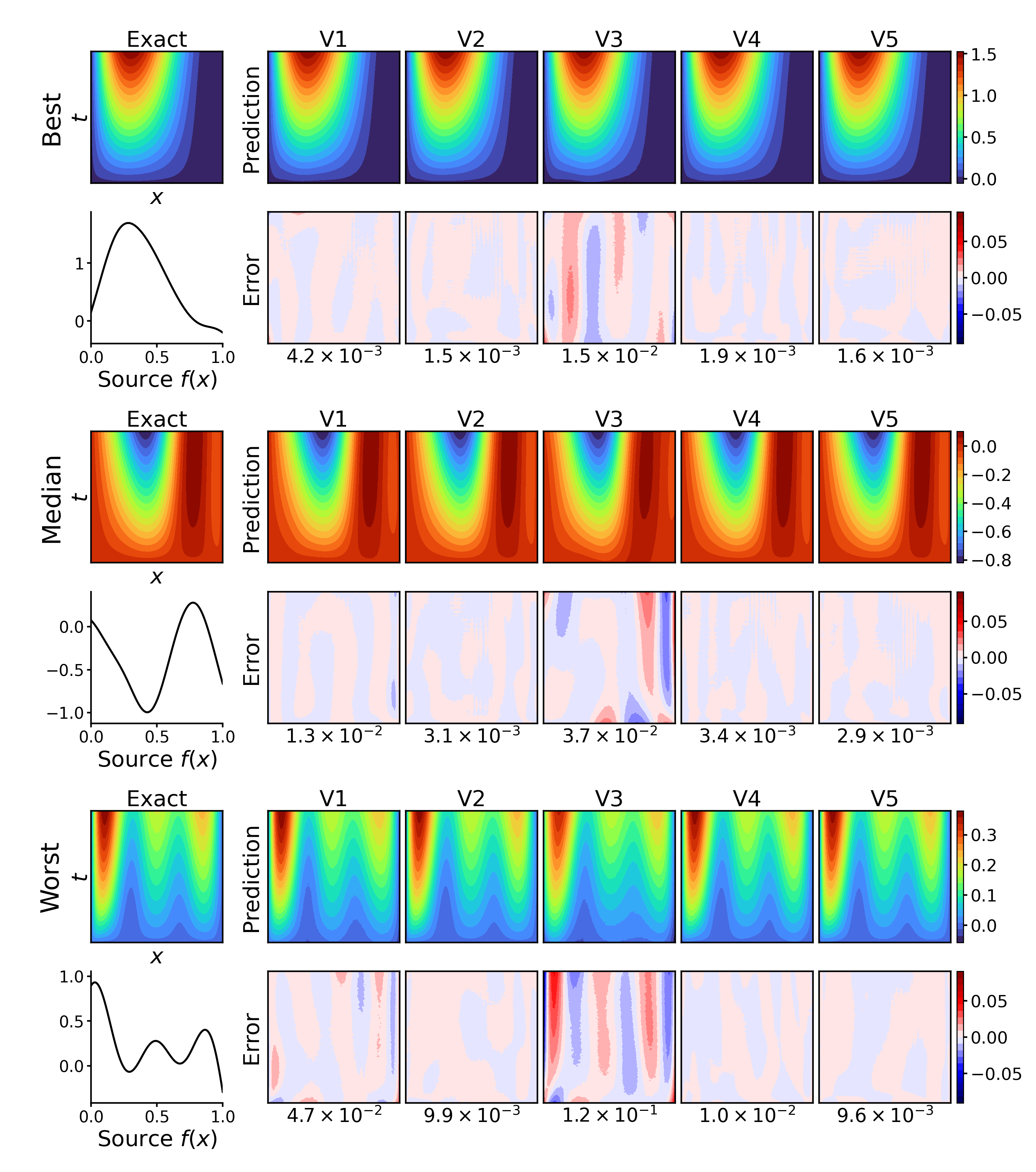}
\caption{Data-driven transient 1D diffusion--reaction. Three held-out source functions, chosen as the
best, median and worst case for the V1 dot-product baseline. In each block the top row is
the exact space-time solution $u(x,t)$ followed by the five variants' predictions, and the
bottom row is the input source $f(x)$ followed by each variant's signed error
(exact $-$ prediction; white is exact). The error color scale is common to all three
blocks; the value beneath each error panel is that variant's relative $L_2$ error on the
case shown.}
\label{fig:dr1d-dd}
\end{figure*}

The representative solution fields for the worst (highest prediction error), median, and best (lowest prediction error) among all test samples for the final time step are shown in Figure~\ref{fig:dr1d-dd} to support these aggregate results.
The cross-attention variants closely reproduce the reference transient space--time fields
and maintain relatively small errors for the best, median, and worst cases of V1
(the dot-product baseline). V3 (self-attention $+$ dot-product fusion) displays broader
and more structured errors, particularly for the median and worst cases. The histogram showing error distribution among test samples for all five architecture  variants in Figure~\ref{fig:app-errhist-dr-dd} in Appendix shows the same separation: the errors
of V5 (self-attention $+$ cross-attention) are concentrated well below those of V1
(the dot-product baseline), whereas the distribution of V3
(self-attention $+$ dot-product fusion) is shifted toward larger errors. Physically,
these reductions show that query-dependent attention better captures how the spatial
source drives the transient field through the combined effects of diffusion and nonlinear
reaction.

\subsubsection{Physics-informed results}
\label{sec:exp:dr:pi}

When training uses only the PDE residual and initial- and boundary-condition losses,
V1 (the dot-product baseline) gives a mean relative $L_2$ error of
$7.64\times10^{-3}$. V4 (global branch $+$ cross-attention) achieves the lowest mean
error, $2.21\times10^{-3}$, followed by V5
(self-attention $+$ cross-attention) at $2.76\times10^{-3}$ and V2
(tokens $+$ cross-attention) at $3.12\times10^{-3}$. V3
(self-attention $+$ dot-product fusion) again performs worse than V1
(the dot-product baseline), with an error of $1.81\times10^{-2}$.

\begin{figure*}[tp]
\centering
\includegraphics[width=0.75\textwidth]{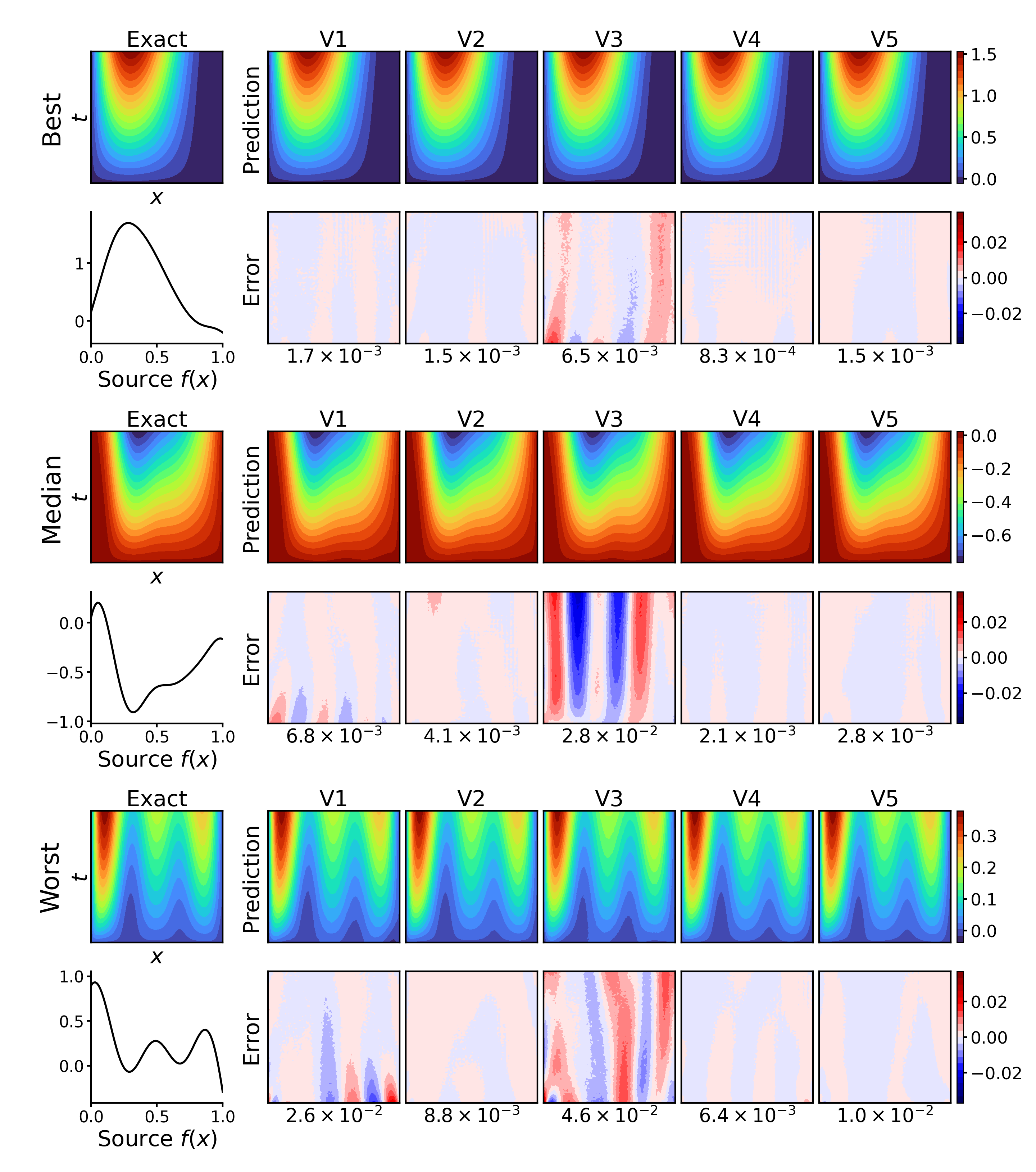}
\caption{Physics-informed transient 1D diffusion--reaction, laid out as in
Fig.~\ref{fig:dr1d-dd}.}
\label{fig:dr1d-pi}
\end{figure*}

Figure~\ref{fig:dr1d-pi} shows that V4 (global branch $+$ cross-attention) produces the
smallest or nearly smallest field errors across the representative cases, including a
substantial reduction for the worst case of V1 (the dot-product baseline).
Figure~\ref{fig:app-errhist-dr-pi} similarly shows a clear leftward shift of the held-out
error distribution of V4 (global branch $+$ cross-attention) relative to V1
(the dot-product baseline). For this comparatively smooth transient diffusion--reaction
problem, every physics-informed variant also achieves a lower mean error than its
corresponding data-driven model. Physically, the result indicates that in this case the physics-informed 
models can well reconstruct the time-dependent response generated by source forcing,
diffusion, and nonlinear reaction while satisfying the homogeneous initial and boundary
conditions.

\FloatBarrier

\subsection{Transient viscous Burgers transport with a variable initial condition}
\label{sec:exp:burgers}

\subsubsection{Problem setup}
\label{sec:exp:burgers:setup}

The operator maps an initial condition to the solution of the transient viscous Burgers equation,
\begin{equation}
\frac{\partial u}{\partial t}
+u\frac{\partial u}{\partial x}
=
\nu\frac{\partial^{2}u}{\partial x^{2}},
\qquad (x,t)\in[0,1]^{2},
\label{eq:burgers}
\end{equation}
with viscosity $\nu=0.01$ and periodic boundary conditions in space:
\begin{equation*}
\begin{aligned}
u(0,t) &= u(1,t),\\
\frac{\partial u}{\partial x}(0,t)
&=
\frac{\partial u}{\partial x}(1,t).
\end{aligned}
\end{equation*}
The input function $a$ is the initial condition $u_{0}(x)=u(x,0)$, drawn from a Gaussian random field with covariance
$\sigma^{2}(-\Delta+\lambda^{2}I)^{-\eta}$ using $\eta=4$, $\lambda=5$, and $\sigma=25$, and sampled at $m=101$ equally spaced sensors. Because the nonlinear term steepens smooth initial data into sharp interior fronts, reference solutions are computed with a spectral exponential-integrator scheme on a fine grid of $4096$ points using the Chebfun library \citep{Driscoll2014}, then recorded on a $101\times101$ grid in $(x,t)$. Of the $2000$ realizations generated, $1600$ are used for training and the remaining $400$ are held out for evaluation, with the split fixed by a stored random seed so that all variants use the same partition.

Training follows the same arrangement as the transient diffusion--reaction benchmark. The data-driven models regress on the reference fields, while the physics-informed models are trained using the residual of Equation~\eqref{eq:burgers}, together with the initial condition and the periodic boundary conditions, following \citet{Wang2021}. Each step samples $100$ initial-condition functions, with $100$ collocation points and $100$ boundary points per function, so the residual and boundary losses are each evaluated at $10^{4}$ space--time points per step.

\subsubsection{Data-driven results}
\label{sec:exp:burgers:dd}

For the data-driven transient Burgers operator, V1 (the dot-product baseline) gives a
mean relative $L_2$ error of $4.06\times10^{-2}$ over 400 held-out initial conditions.
V5 (self-attention $+$ cross-attention) and V4
(global branch $+$ cross-attention) obtain nearly identical mean errors of
$5.27\times10^{-3}$ and $5.36\times10^{-3}$, respectively, while V2
(tokens $+$ cross-attention) reaches $7.53\times10^{-3}$.
V3 (self-attention $+$ dot-product fusion) does not
improve upon V1 (the dot-product baseline) and gives a mean error of
$4.59\times10^{-2}$.

\begin{figure*}[tp]
\centering
\includegraphics[width=0.75\textwidth]{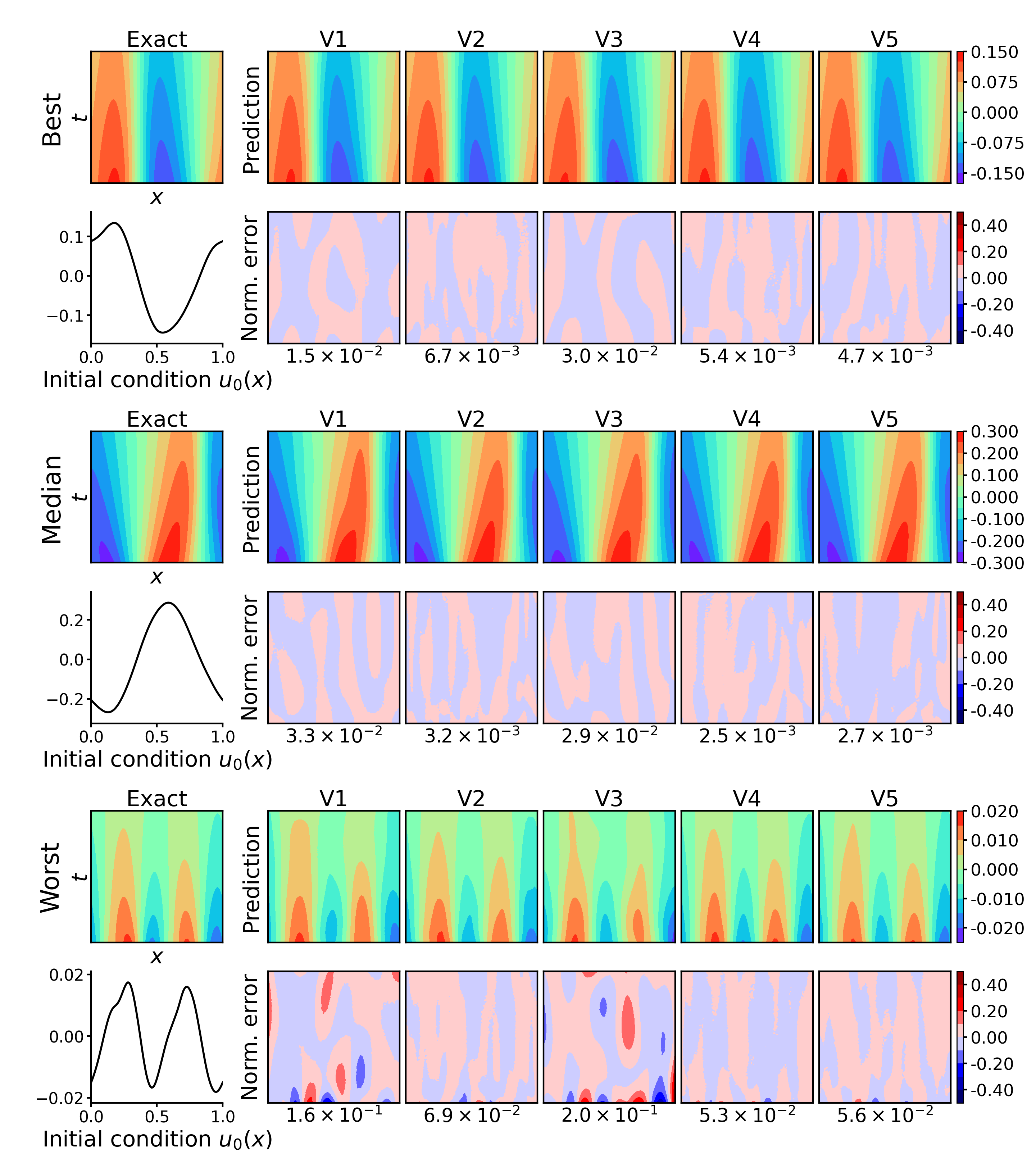}
\caption{Data-driven transient 1D Burgers. Three held-out initial conditions, chosen as the best,
median and worst case for the V1 dot-product baseline. In each block the top row is the
reference space-time solution $u(x,t)$ followed by the five variants' predictions, and the
bottom row is the initial condition $u_0(x)$ followed by each variant's normalised signed
error $(\mathrm{exact}-\mathrm{prediction})/\max|\mathrm{exact}|$, where the normalization (Norm)
is by the peak magnitude of the exact field for that case; white is exact. The value beneath each error panel is
that variant's relative $L_2$ error on the case shown.}
\label{fig:burgers1d-dd}
\end{figure*}

As illustrated in Figure~\ref{fig:burgers1d-dd} for the best, median, and worst V1 test cases, V4
(global branch $+$ cross-attention) and V5
(self-attention $+$ cross-attention) more accurately follow the transport and steepening
of the transient solution features generated from the initial condition. The difference
is most visible in the difficult cases of V1 (the dot-product baseline), where the
dot-product models produce larger errors around the evolving fronts.
Figure~\ref{fig:app-errhist-burgers-dd} shows the error distributions over all test samples for all five network architectures and confirms that the improvement is consistent over
the entire test set, with the distribution of V5 (self-attention $+$ cross-attention)
concentrated well below that of V1 (the dot-product baseline). Physically, this indicates that cross-attention captures the interaction between nonlinear advection and viscous diffusion more effectively than the classical dot-product fusion as the initial profile evolves into translating and steepening fronts.

\subsubsection{Physics-informed results}
\label{sec:exp:burgers:pi}

The physics-informed transient Burgers problem is substantially more challenging for the dot-product
architectures. V1 (the dot-product baseline) gives a mean relative $L_2$ error of
$1.24\times10^{-1}$, and V3 (self-attention $+$ dot-product fusion) increases the error
to $1.92\times10^{-1}$. By comparison, V4
(global branch $+$ cross-attention) reduces the mean error to $7.04\times10^{-3}$,
the lowest of the five variants. V5
(self-attention $+$ cross-attention) and V2 (tokens $+$ cross-attention) reach
$1.03\times10^{-2}$ and $1.58\times10^{-2}$, respectively.

\begin{figure*}[tp]
\centering
\includegraphics[width=0.75\textwidth]{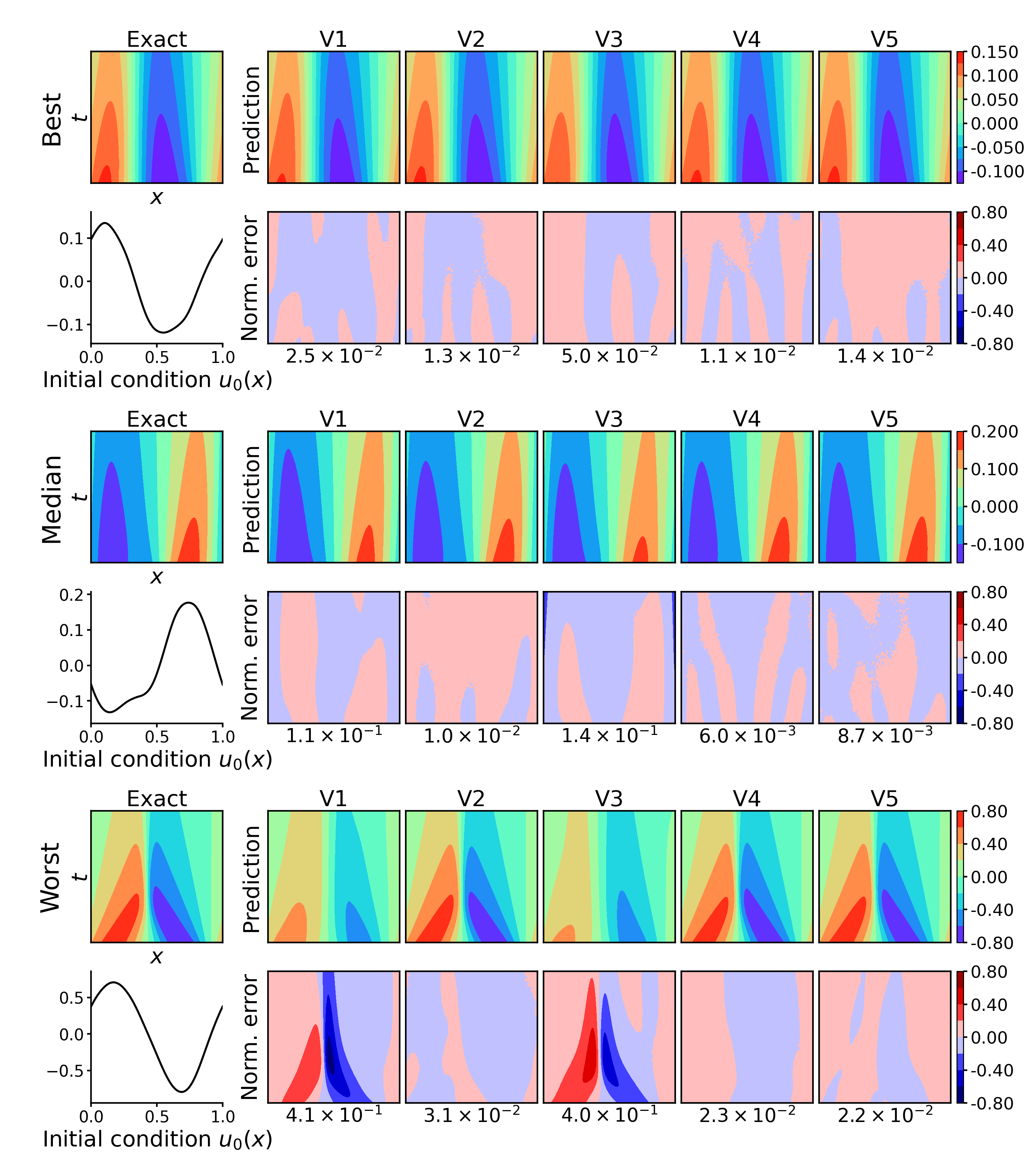}
\caption{Physics-informed transient 1D Burgers, laid out as in Fig.~\ref{fig:burgers1d-dd}.}
\label{fig:burgers1d-pi}
\end{figure*}

Figure~\ref{fig:burgers1d-pi}, and particularly its normalized error contour rows, shows that the cross-attention models recover the location
and shape of the transient advected fronts much more accurately than V1
(the dot-product baseline) and V3 (self-attention $+$ dot-product fusion). For the worst
case of V1 (the dot-product baseline), the relative error decreases from approximately
$4.1\times10^{-1}$ for V1 (the dot-product baseline) to approximately
$2.3\times10^{-2}$ and $2.2\times10^{-2}$ for V4
(global branch $+$ cross-attention) and V5
(self-attention $+$ cross-attention). The error distributions in
Figure~\ref{fig:app-errhist-burgers-pi} show a large corresponding separation between the best
cross-attention model and the baseline across the 400 test functions. Physically, the
improvement shows that attention-based fusion considerably helps physics-informed training resolve the
competition between nonlinear advection and viscosity, especially near rapidly evolving
fronts where spatial derivatives are largest.

\FloatBarrier

\subsection{Poisson heat conduction with a heterogeneous source}
\label{sec:exp:poisson}

\subsubsection{Problem setup}
\label{sec:exp:poisson:setup}

The operator maps a heat source to the steady temperature field governed by the Poisson
equation,
\begin{equation}
\kappa \nabla^{2} u + f(x,y) = 0,
\qquad (x,y)\in[0,1]^{2},
\qquad u = 0 \ \text{ on } \partial\Omega,
\label{eq:poisson}
\end{equation}
with conductivity $\kappa = 0.01$. The input function $a$ is the heat source $f$. This
extends the parametric heat-source setting of
\citet{KoricAbueidda2023} to a rougher and more varied family of sources. Each source is a
Gaussian random field with a power-law spectrum $|\xi|^{-\alpha}$ in the wavevector
$\xi$, with $\alpha = 3.2$,
synthesised in Fourier space following \citet{Sciolla2017}, then given a random amplitude
and a random constant offset so that the fields differ in magnitude and mean rather than
in shape alone. Sources and solutions are represented on a $32\times32$ grid that includes
the boundary, giving $m = 1024$ sensors, an order of magnitude more sensors than either
one-dimensional benchmark. Reference solutions are obtained by inverting the five-point
Dirichlet Laplacian exactly with a discrete sine transform, so the reference satisfies the
discretized equation to machine precision. We generate $5000$ sources for training and
$1000$ held-out (test) sources for evaluation, again from a separate random stream.

The physics-informed models are trained from the PDE residual of Equation~\eqref{eq:poisson}
at interior collocation points, using second derivatives of the prediction with respect to
the query coordinates, together with a term enforcing $u=0$ on the boundary; the two terms
are weighted equally. Each step uses $64$ source distributions, each with $32$ interior and $32$ boundary
points, giving $2048$ points per loss term per step. Reference solutions enter only through the reported errors.

\subsubsection{Data-driven results}
\label{sec:exp:poisson:dd}

The largest differences among the architectures occur for the data-driven Poisson
problem. V1 (the dot-product baseline) gives a mean relative $L_2$ error of
$2.95\times10^{-1}$, whereas V5 (self-attention $+$ cross-attention) reduces it to
$9.13\times10^{-3}$. V2 (tokens $+$ cross-attention) and V4
(global branch $+$ cross-attention) also perform well, with errors of
$1.05\times10^{-2}$ and $1.43\times10^{-2}$, respectively. Unlike the transient
one-dimensional benchmarks, V3 (self-attention $+$ dot-product fusion) also improves
substantially over V1 (the dot-product baseline), reaching $3.81\times10^{-2}$.

\begin{figure*}[tp]
\centering
\includegraphics[width=0.75\textwidth]{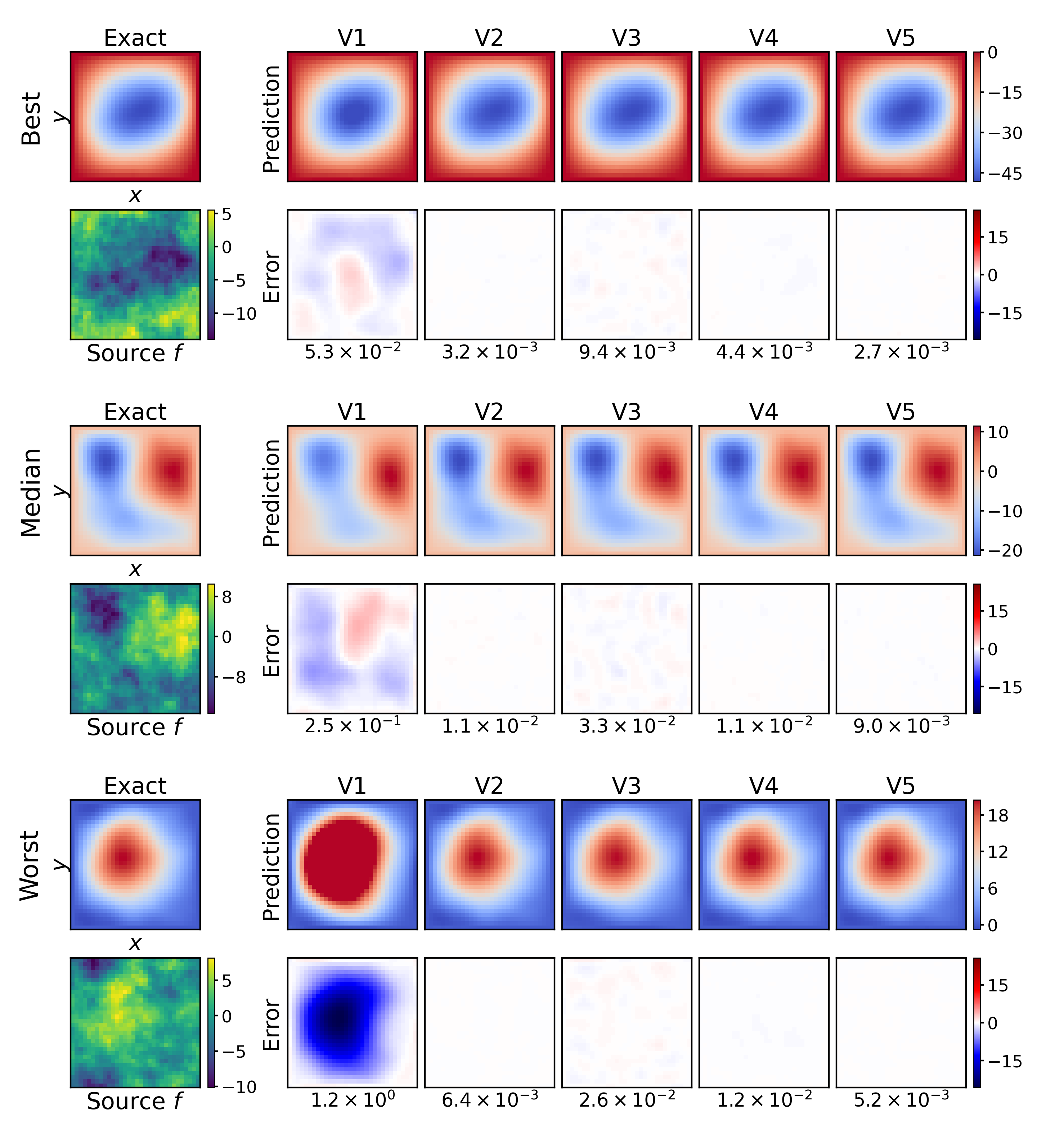}
\caption{Data-driven 2D Poisson heat conduction. Three held-out source fields, chosen as
the best, median and worst case for the V1 dot-product baseline. In each block the top row
is the reference solution $u(x,y)$ followed by the five variants' predictions, and the
bottom row is the input source $f(x,y)$ followed by each variant's signed error
(exact $-$ prediction; white is exact). The error colour scale is common to all three
blocks; the value beneath each error panel is that variant's relative $L_2$ error on the
case shown.}
\label{fig:poisson-dd}
\end{figure*}

The solution field comparisons in Figure~\ref{fig:poisson-dd} make this difference particularly
clear. The prediction of V1 (the dot-product baseline) can contain large errors in
solution magnitude and spatial structure. This is particularly visible for its worst test case, whereas the
attention variants accurately reproduce the reference temperature field. In that case, the
error of V1 (the dot-product baseline) is approximately $1.2$, compared with
$5.2\times10^{-3}$ for V5 (self-attention $+$ cross-attention). The error 
distributions for test samples in Figure~\ref{fig:app-errhist-poisson-dd} show that this improvement
extends across the full set of 1000 source fields. Physically, the result shows that
attention better represents the nonlocal elliptic mapping from a heterogeneous heat source
to the steady, diffusion-governed temperature field in the interior while recovering the
zero-valued Dirichlet boundary condition.

\subsubsection{Physics-informed results}
\label{sec:exp:poisson:pi}

For physics-informed Poisson, V1 (the dot-product baseline) gives a mean relative $L_2$
error of $3.34\times10^{-1}$. V5 (self-attention $+$ cross-attention) achieves the
the lowest mean error, $2.88\times10^{-2}$. V3 (self-attention $+$ dot-product fusion) ranks second by mean error at
$3.96\times10^{-2}$, followed by V2 (tokens $+$ cross-attention) at
$6.15\times10^{-2}$ and V4 (global branch $+$ cross-attention) at
$7.04\times10^{-2}$. The comparatively strong performance of both V3
(self-attention $+$ dot-product fusion) and V5
(self-attention $+$ cross-attention) distinguishes Poisson from the two transient
one-dimensional benchmarks and shows the value of branch self-attention for the larger
two-dimensional input field.

\begin{figure*}[tp]
\centering
\includegraphics[width=0.75\textwidth]{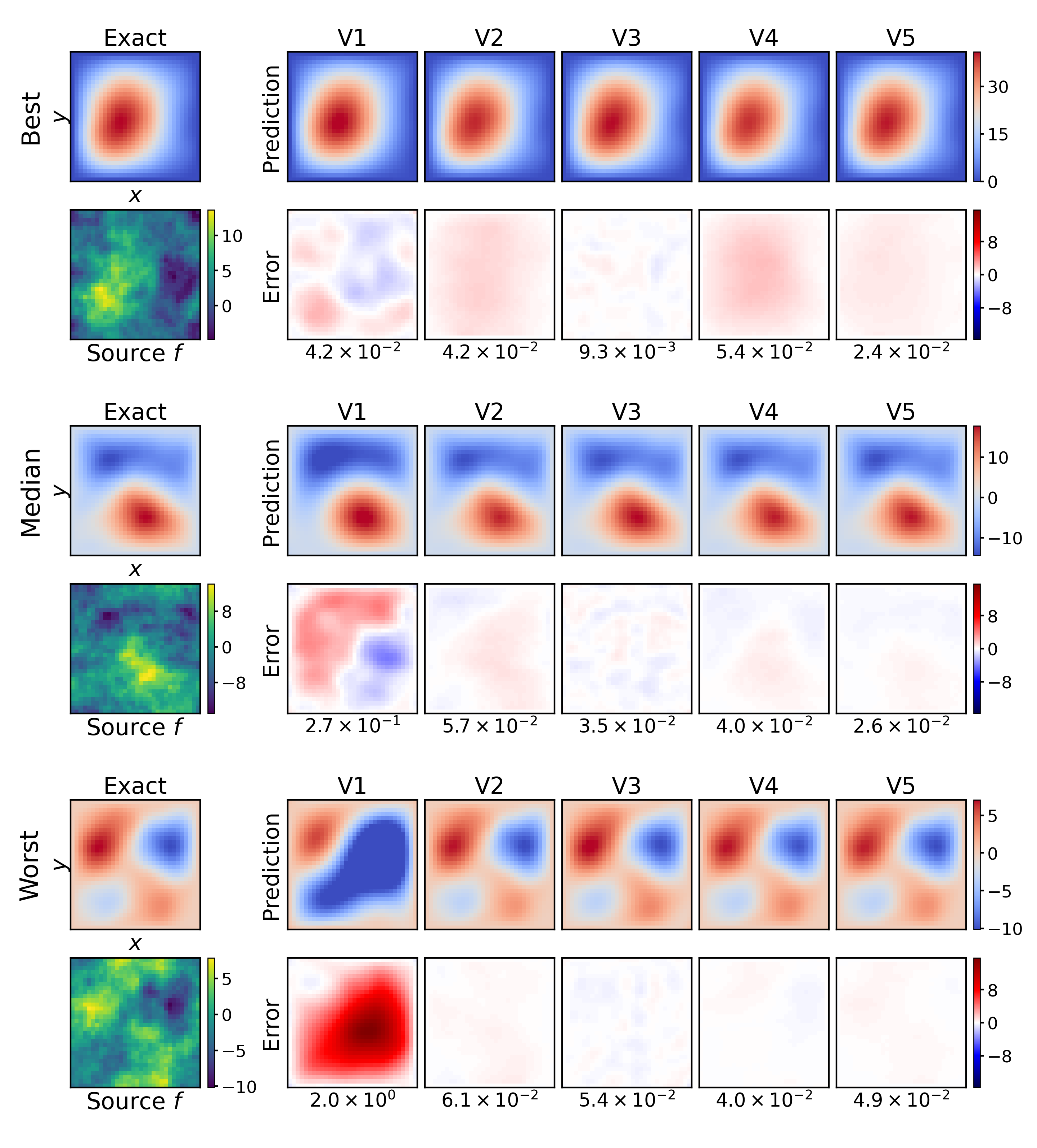}
\caption{Physics-informed 2D Poisson heat conduction, laid out as in
Fig.~\ref{fig:poisson-dd}.}
\label{fig:poisson-pi}
\end{figure*}

Figure~\ref{fig:poisson-pi} shows that V5
(self-attention $+$ cross-attention), V3 (self-attention $+$ dot-product fusion), and the
other attention variants substantially reduce the broad amplitude and spatial errors
visible in the predictions of V1 (the dot-product baseline).
Figure~\ref{fig:app-errhist-poisson-pi} confirms that the error distribution of V5
(self-attention $+$ cross-attention) is shifted well below that of V1
(the dot-product baseline) over the 1000 held-out source fields. Physically, these results show that, compared with the classical dot-product formulation, physics-informed models enhanced with attention mechanisms can notably more accurately learn the global diffusive heat-conduction response driven by challenging spatial heat-source distributions without using reference solution fields.

\FloatBarrier

\subsection{Cross-attention depth}
\label{sec:exp:depth}
The preceding results identify the fusion operator as the main source of the accuracy
gain, and the transformer decoder of \citet{Vaswani2017} stacks several cross-attention
layers rather than one. It is therefore worth asking whether additional depth in the
fusion buys further accuracy, and whether the gain justifies its cost.

The sweep is run on V2 (tokens $+$ cross-attention), which combines per-sensor
tokenization with a single cross-attention operation, using the transient
diffusion--reaction benchmark under both training paradigms. The single operation is
replaced by a stack of $N$ pre-normalised (LN) blocks, the
$\ell$-th of which refines the query as
\begin{equation}
Q^{(\ell)}
=
Q^{(\ell-1)}
+
\mathrm{Attn}\big(\mathrm{LN}(Q^{(\ell-1)}),\,K,\,V\big),
\qquad \ell = 1,\dots,N,
\label{eq:ca-stack}
\end{equation}
starting from the trunk representation $Q^{(0)}$ of Equation~\eqref{eq:qkv}. The branch
tokens are encoded once and shared by every block, so only the query is updated as it
passes through the stack. Nothing else changes with $N$, which isolates depth from the
block type and from the branch encoding. We take $N=1,2,4,8$, with $N=1$ the single-block
fusion used in the earlier parts of the study.

Table~\ref{tbl:depth} collects the outcome. Mean error falls monotonically with
$N$ under both paradigms, but not in proportion to the cost: going from $N=1$ to $N=8$
lowers the mean error by a factor of $2.1$ under data-driven training and $1.8$ under
physics-informed training, while parameters grow $2.9\times$ and inference time
$3.2\times$. Where the two paradigms differ is in what depth costs during training. The
data-driven models are almost unaffected up to $N=4$, which reduces the mean error by
$31.6\%$ for a $5.2\%$ increase in training time. The physics-informed models pay for the
same depth: $N=4$ reduces the mean error by $38.4\%$ but takes $2.49$ times as long to
train, since the residual requires second derivatives to be propagated through every added
block. Beyond $N=4$ the physics-informed gain narrows sharply, to a further $10.9\%$ for
another $54\%$ of training time. At $N=8$ the two paradigms arrive at nearly the same
error, $1.626\times10^{-3}$ and $1.633\times10^{-3}$, from very different training budgets.
A single block therefore captures most of the available benefit, which is what we adopt
elsewhere in this study.

\begin{table}[width=.95\linewidth,cols=5,pos=h]
\caption{Effect of stacking $N$ cross-attention blocks on the transient diffusion--reaction
benchmark, under both training paradigms. Each error entry is the mean relative $L_2$ error
over the $100$ held-out source functions $\pm$ one standard deviation across those
functions.}
\label{tbl:depth}
\small
\begin{tabular*}{\tblwidth}{@{\extracolsep{\fill}}lrrrr@{}}
\toprule
 & $N=1$ & $N=2$ & $N=4$ & $N=8$ \\
\midrule
Parameters              & 182{,}401 & 231{,}809 & 330{,}625 & 528{,}257 \\
Inference (ms)          & 3.05 & 3.97 & 5.82 & 9.62 \\
\midrule
\multicolumn{5}{l}{\textit{Data-driven}} \\
Mean $L_2$ ($\times 10^{-3}$)        & $3.343 \pm 1.796$ & $2.360 \pm 1.072$ & $2.287 \pm 1.054$ & $1.626 \pm 0.698$ \\
Training time           & 4m 10s & 4m 02s & 4m 23s & 6m 52s \\
\midrule
\multicolumn{5}{l}{\textit{Physics-informed}} \\
Mean $L_2$ ($\times 10^{-3}$)        & $2.976 \pm 1.373$ & $2.286 \pm 1.009$ & $1.832 \pm 0.845$ & $1.633 \pm 0.692$ \\
Training time           & 25m 37s & 35m 06s & 63m 43s & 98m 00s \\
\bottomrule
\end{tabular*}
\end{table}

\subsection{Computational cost}
\label{sec:exp:cost}
All training and inference measurements in Table~\ref{tbl:cost} were obtained on the DeltaAI computing system hosted at the National Center for Supercomputing Applications (NCSA) using a single NVIDIA GH200 GPU equipped with 96~GB of device memory. Training cost increases more steeply with architectural complexity under physics-informed training. Across the benchmark-specific measurements, the median ratio of physics-informed to data-driven training time is $2.87\times$ for the dot-product variants V1 and V3, compared with $7.82\times$ for the cross-attention variants V2, V4, and V5. Averaged across the three benchmarks, replacing V1 with the complete V5 attention architecture increases training time from 1~min 37~s to 7~min 46~s for data-driven learning, a factor of $4.82\times$, and from 4~min 28~s to 28~min 28~s for physics-informed learning, a factor of $6.37\times$. The data-driven Poisson measurements for V3 and V5 are notable outliers because branch self-attention operates over 1024 sensor tokens, compared with approximately 100 tokens in the one-dimensional benchmarks, and its computational complexity is quadratic in the token count. Inference cost is determined primarily by the architecture and is nearly unchanged between the two training paradigms. V1 is the fastest variant for the one-dimensional benchmarks, while V3 and V5 incur the highest inference times because they apply self-attention over the sensor tokens. The similar costs of V2 and V4 indicate that injecting global branch context adds little inference overhead beyond cross-attentive fusion. Despite these differences, all models evaluate a complete solution field in less than $12$ ms, preserving the rapid online evaluation expected of neural operators, which are several orders of magnitude faster than classical numerical PDE solvers.

\begin{table*}[width=\textwidth,cols=7,pos=t]
\caption{Training wall-clock time and inference time for a full-field forward pass, on a single NVIDIA H200 GPU. Panel A: training. Panel B: inference.}
\label{tbl:cost}
\small
\begin{tabular*}{\tblwidth}{@{\extracolsep{\fill}}lrrrrrr@{}}
\toprule
 & \multicolumn{3}{c}{Data-driven} & \multicolumn{3}{c}{Physics-informed} \\
\cmidrule(lr){2-4}\cmidrule(lr){5-7}
Variant & Tr. diff.-react. & Tr. Burgers & Poisson & Tr. diff.-react. & Tr. Burgers & Poisson \\
\midrule
\multicolumn{7}{l}{\textit{Panel A: training wall-clock time}} \\
V1 dot product   & 1m 49s & 1m 50s & 1m 11s  & 5m 11s  & 5m 59s  & 2m 14s  \\
V2 tokens $+$ CA & 2m 57s & 2m 43s & 3m 01s  & 31m 21s & 27m 49s & 16m 07s \\
V3 SA $+$ dot    & 2m 18s & 2m 20s & 15m 14s & 6m 39s  & 7m 18s  & 16m 16s \\
V4 global $+$ CA & 3m 18s & 3m 20s & 3m 08s  & 23m 57s & 28m 51s & 16m 15s \\
V5 SA $+$ CA     & 3m 17s & 3m 09s & 16m 51s & 25m 41s & 29m 59s & 29m 45s \\
\midrule
\multicolumn{7}{l}{\textit{Panel B: inference time (ms)}} \\
V1 dot product   & 0.18  & 0.50  & 0.540 & 0.17  & 0.50  & 0.370 \\
V2 tokens $+$ CA & 3.07  & 3.48  & 0.399 & 3.08  & 3.48  & 0.400 \\
V3 SA $+$ dot    & 8.86  & 10.46 & 0.363 & 8.73  & 10.53 & 0.384 \\
V4 global $+$ CA & 3.14  & 3.54  & 0.390 & 3.14  & 3.53  & 0.418 \\
V5 SA $+$ CA     & 10.02 & 11.76 & 0.411 & 10.00 & 11.69 & 0.446 \\
\bottomrule
\end{tabular*}
\end{table*}

\FloatBarrier

\section{Conclusions, Limitations and Future Work}
\label{sec:conclusions}

This work presents a systematic study of attention mechanisms in deep operator
networks across both data-driven and physics-informed training regimes. Rather
than comparing complete architectures that differ in many ways, the study
isolated the effects of branch--trunk fusion, branch self-attention, global
pre-mixing, and cross-attention depth.

The central finding is that query-dependent cross-attention is the most
consistent source of improvement. Relative to the classical dot-product V1,
per-sensor tokenization with cross-attention, V2, reduced the mean relative
$L_2$ error in all six benchmark--training combinations by factors ranging
from $2.4$ to $28.0$. The best attention configuration improved upon the
baseline by $4.3\times$, $7.7\times$, and $32.3\times$ for the three
data-driven problems and by $3.5\times$, $17.6\times$, and $11.6\times$ for
their physics-informed counterparts. The classical dot product combines one
global branch vector with the query representation in the same fixed way at
every output location. Cross-attention instead retains the input as spatially
identified tokens and allows each query coordinate to assign different weights
to those tokens. It can therefore adapt the input--solution interaction to the
output location.

The role of self-attention in branch is more conditional. The self-attentive
token branch followed by dot-product fusion in V3 is worse than the classical
baseline V1 in all four one-dimensional benchmark--training combinations. Its
mean error is $2.9\times$ and $2.4\times$ larger than V1 for data-driven and
physics-informed diffusion--reaction, and is also larger for both Burgers
cases. Self-attention can organize relationships among input tokens in those
cases, but it cannot remove the fixed branch--trunk fusion bottleneck. In
contrast, V3 improves upon V1 by $7.7\times$ and $8.4\times$ for data-driven
and physics-informed Poisson. In this case, self-attention can better represent
nonlocal spatial relationships among 1024 tokens from a complex two-dimensional
source field. What separates the two regimes is the input function rather than
the mechanism, and the effect changes sign rather than merely magnitude. Branch
self-attention pays only once the sampled input is large and rough enough that
pooling it into a single vector would discard structure, as with the $1024$
sensors of the Poisson source. Below that threshold, as with the roughly one
hundred smoothly varying sensors of the one-dimensional benchmarks, it is
actively harmful rather than simply redundant.

Self-attention becomes more reliable and consistent when combined with
cross-attention in V5. Comparing V5 with V2 shows that adding branch
self-attention improves all six cases, although the improvement is smaller than
that obtained by changing the fusion mechanism. The gain ranges from
approximately $1.09\times$ for data-driven diffusion--reaction to $2.13\times$
for physics-informed Poisson. Thus, self-attention is most useful as an
input-field encoder that prepares spatially connected tokens for subsequent
query-dependent fusion. It should be considered for large, multidimensional, or
highly heterogeneous input functions, but it is not a substitute for
cross-attention.

In V4, adding a globally mixed branch representation to the sensor tokens before
cross-attention fusion does not consistently provide an advantage either. It
improves V4 over V2 for Burgers under both training paradigms and yields the
best physics-informed results for diffusion--reaction and Burgers. However, V4
is less accurate than V2 for data-driven diffusion--reaction and for both
Poisson cases. For a heterogeneous two-dimensional source, global pre-mixing may
compress or blur spatial detail already available to cross-attention through
individual tokens. V4 also has the largest parameter count but is not
consistently the most accurate model. Global pre-mixing should therefore be
treated as a problem-dependent option rather than a default architectural
component.

Physics-informed operator learning with cross-attention is more numerically complex and computationally expensive than its
data-driven counterpart because automatic differentiation must propagate the
PDE's spatial and temporal derivatives through the coordinate-dependent queries,
attention weights, softmax operations, and stacked attention blocks, whereas
in data-driven training, the predicted output is only evaluated, not differentiated via attention.  That penalty is not shared equally by the architectures. Moving from
data-driven to physics-informed training multiplies the median training time by
$2.87\times$ for the dot-product variants but by $7.82\times$ for the
cross-attention variants, so attention is roughly three times more expensive to
train under physics-informed learning than under a data-driven counterpart. However, unlike data-driven models, physics-informed learning does not require time-consuming label generation for training data from classical numerical analysis or experiments. The physics-informed models are generally less accurate for Burgers
and Poisson, where nonlinear front formation or rough two-dimensional inputs along steeper localized spatial derivatives make residual-based optimization more demanding.

The cross-attention depth study within V2 raises the parameter count by
$2.9\times$ for only a $2.1\times$ reduction in error (Table~\ref{tbl:depth}),
whereas the $1.6\times$ parameter increase from V1 to V2 reduces the error by
factors of 2.4--28.0. This shows that capacity alone cannot account for the
observed gains.
Attention also increases inference cost, but only slightly at this scale: every
trained model evaluates a complete solution field in under $12$ ms for a new set of input functions without any additional training and orders of magnitude faster than repeated PDE solutions obtained with
conventional numerical solvers.

The study considers three PDEs on fixed, regular one- and two-dimensional
domains with several limitations left for future work, such as three-dimensional
irregular domains, systematic comparison of positional encodings, alternative designs with
separate encoders for sensor coordinates and function values, as used for keys
and values in PINTO \citep{BoyaSubramani2025}, and studying more advanced
neural-operator-transformer architectures specialized in sequential
\citep{Liu2026a} and geometry-dependent \citep{Liu2026b} operator learning. Fourier neural operators lack the separate branch and trunk representations of DeepONet, so they offer no fusion step for cross-attention to act on, but the remaining mechanisms studied here, in particular self-attention over the input field, could be examined in the future work in the similar controlled manner within the FNO family.
\section*{Acknowledgments.}
    This research utilized the Delta and DeltaAI advanced computing and data resources,
    which are supported by the National Science Foundation (awards OAC-2005572 and
    OAC-2320345) and the State of Illinois. Delta and DeltaAI are joint efforts of
    the University of Illinois Urbana-Champaign and its National Center for Supercomputing
    Applications (NCSA). This work was in part supported by the NSF grant
    2209875. The authors would also like to thank NCSA at the University of
    Illinois, particularly its Research Directorate, Industry Program,
    and the Center for Artificial Intelligence Innovation (CAII), for their
    support.

    \section*{Disclosure of Interests.}
    The authors have no competing interests to declare that are relevant to the
    content of this article.

    \section*{Replication of Results.}
    The data and source code supporting this study will be made available in a
    public GitHub repository upon acceptance of the paper.
    
\section*{Appendix}
\appendix

\section{Held-out (Test) error distributions}
\label{app:errhist}

\begin{figure}[pos=h!]
\centering
\includegraphics[width=0.74\textwidth]{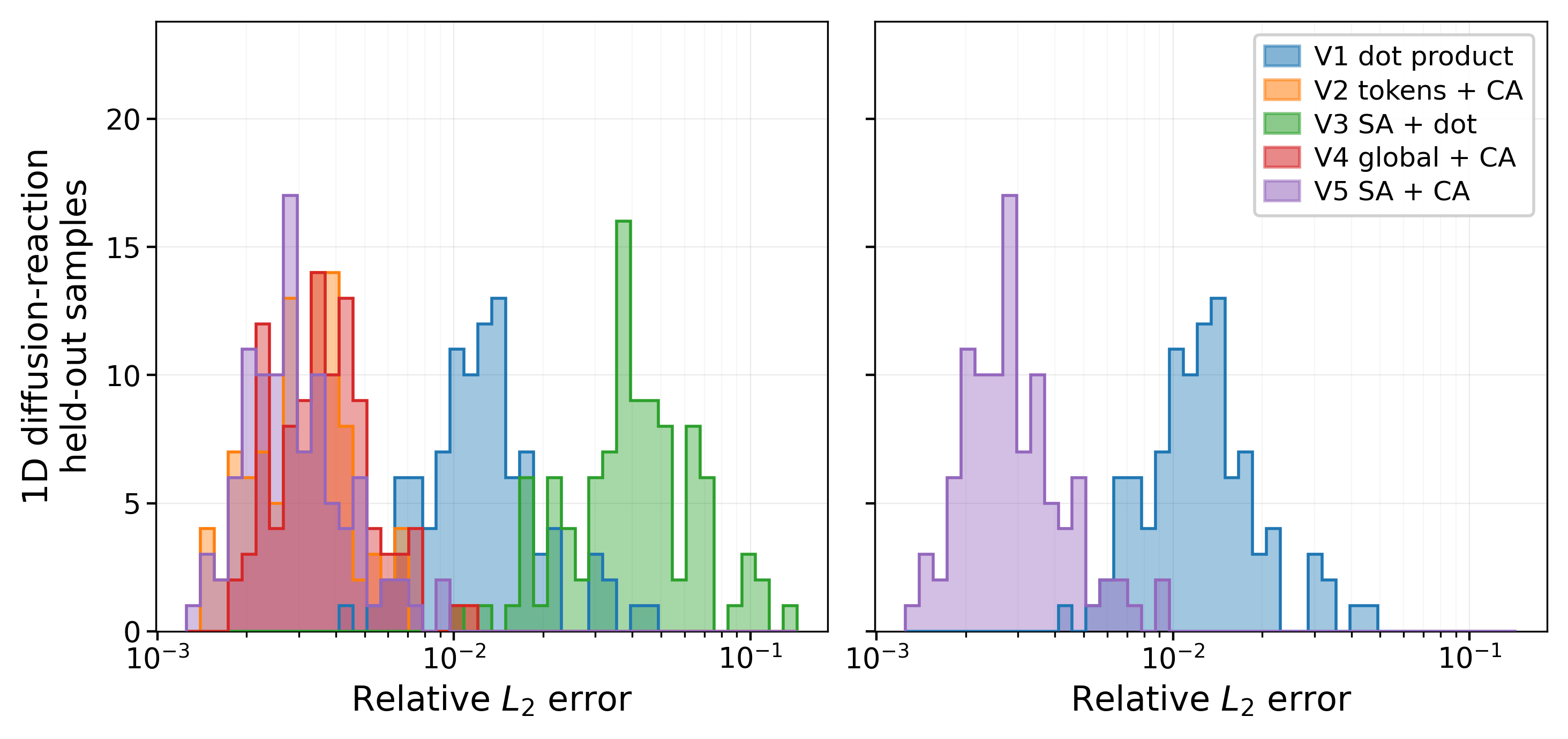}
\caption{Held-out error distributions for the data-driven diffusion-reaction models over
100 matched test samples. Left: all five variants. Right: the V1 dot-product
baseline, mean relative $L_2$ error $1.38\times10^{-2}$, against the best variant
V5 at $3.19\times10^{-3}$.}
\label{fig:app-errhist-dr-dd}
\end{figure}

\begin{figure}[pos=h!]
\centering
\includegraphics[width=0.74\textwidth]{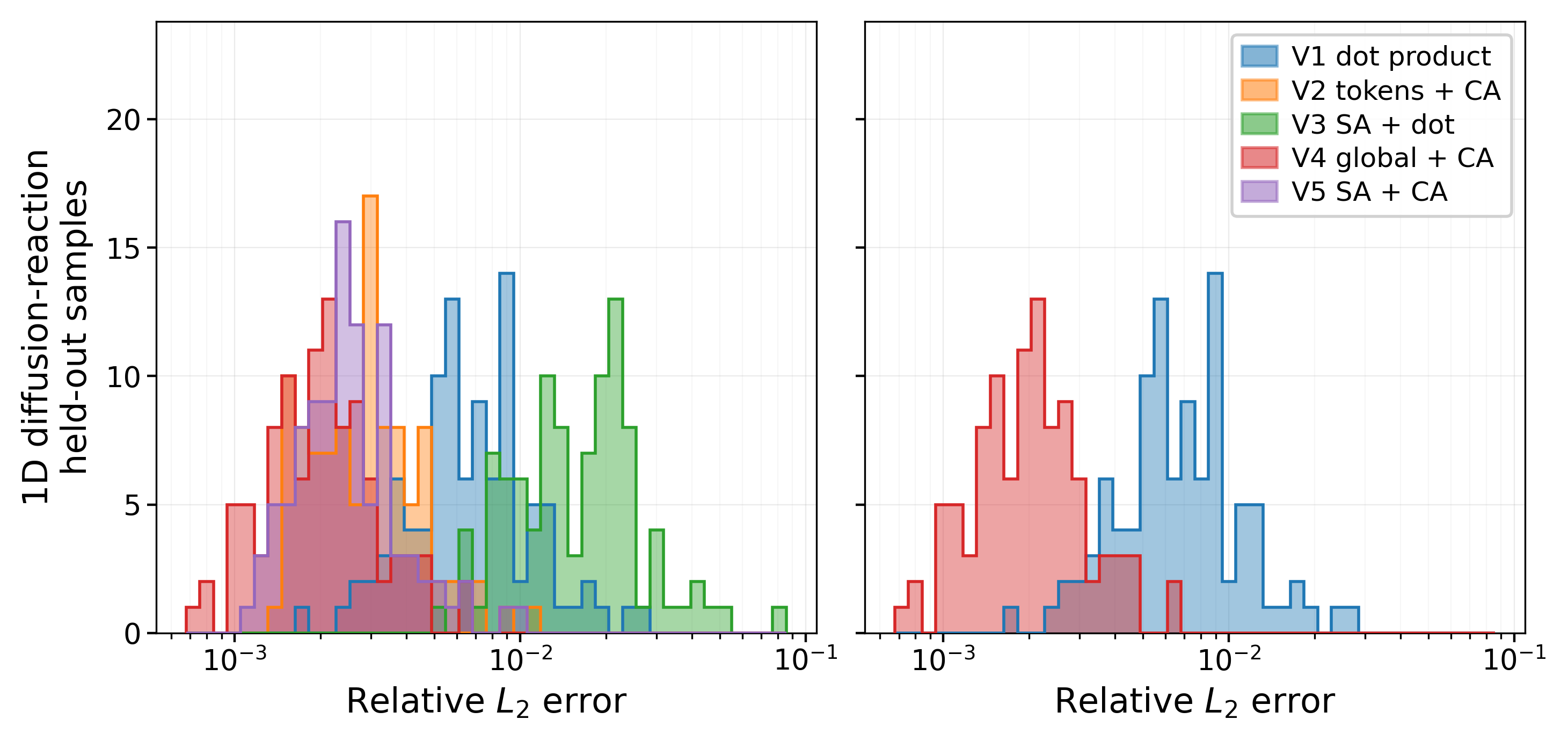}
\caption{Held-out error distributions for the physics-informed diffusion-reaction models
over 100 matched test samples. Left: all five variants. Right: the V1 dot-product
baseline, mean relative $L_2$ error $7.64\times10^{-3}$, against the best variant
V4 at $2.21\times10^{-3}$.}
\label{fig:app-errhist-dr-pi}
\end{figure}

\begin{figure}[pos=h!]
\centering
\includegraphics[width=0.74\textwidth]{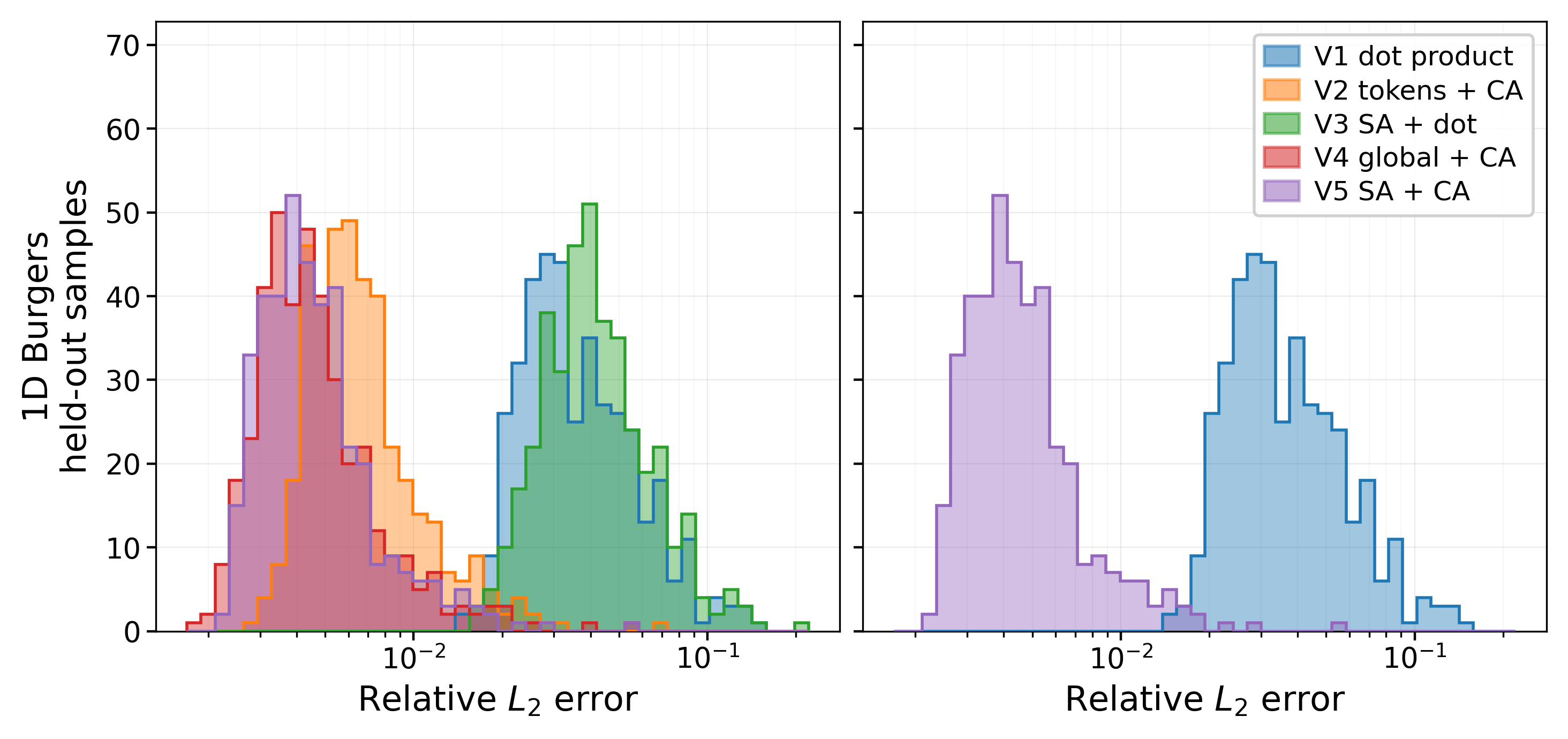}
\caption{Held-out error distributions for the data-driven Burgers models over 400
matched test samples. Left: all five variants. Right: the V1 dot-product
baseline, mean relative $L_2$ error $4.06\times10^{-2}$, against the best variant
V5 at $5.27\times10^{-3}$.}
\label{fig:app-errhist-burgers-dd}
\end{figure}

\begin{figure}[pos=h!]
\centering
\includegraphics[width=0.74\textwidth]{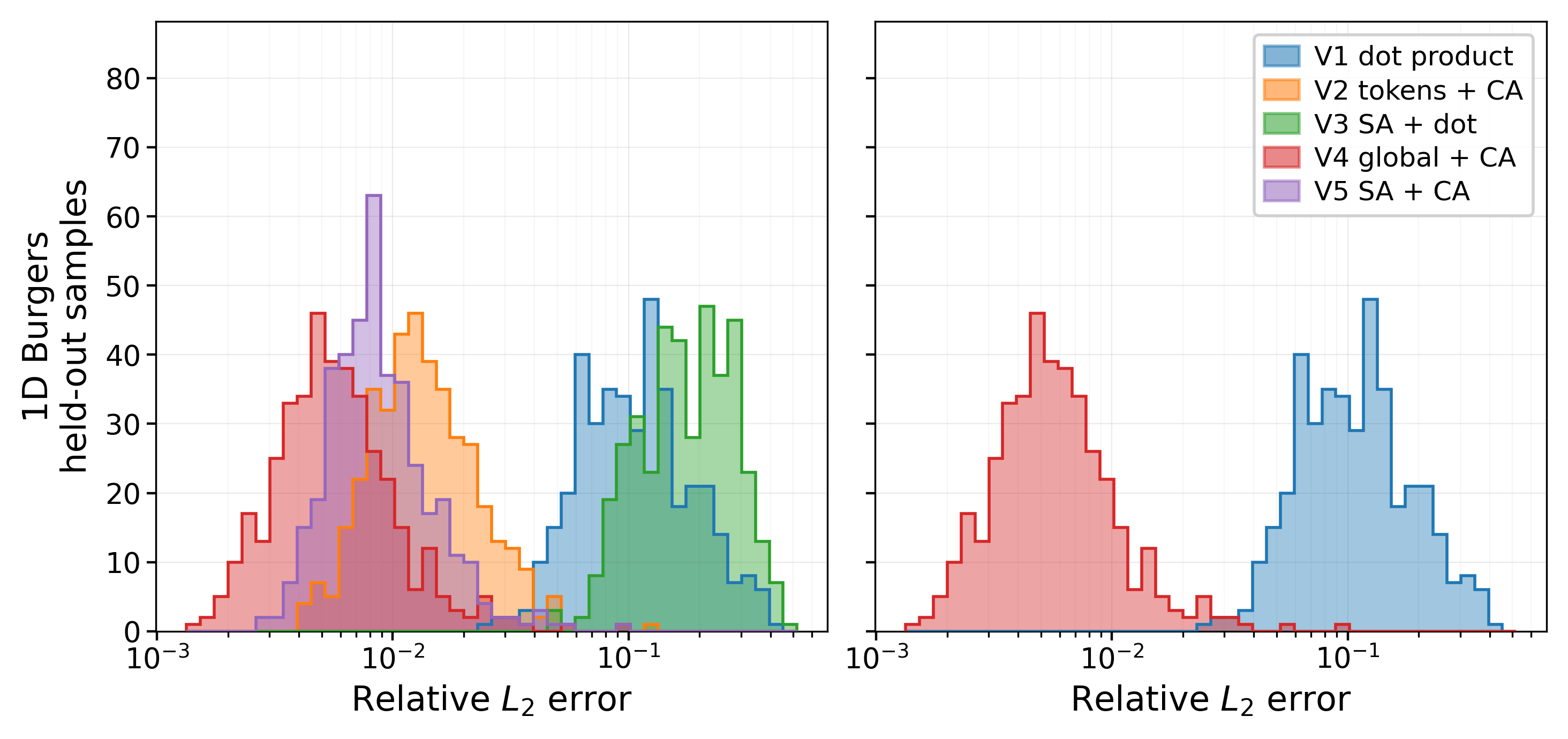}
\caption{Held-out error distributions for the physics-informed Burgers models over 400
matched test samples. Left: all five variants. Right: the V1 dot-product
baseline, mean relative $L_2$ error $1.24\times10^{-1}$, against the best variant
V4 at $7.04\times10^{-3}$.}
\label{fig:app-errhist-burgers-pi}
\end{figure}

\begin{figure}[pos=h!]
\centering
\includegraphics[width=0.74\textwidth]{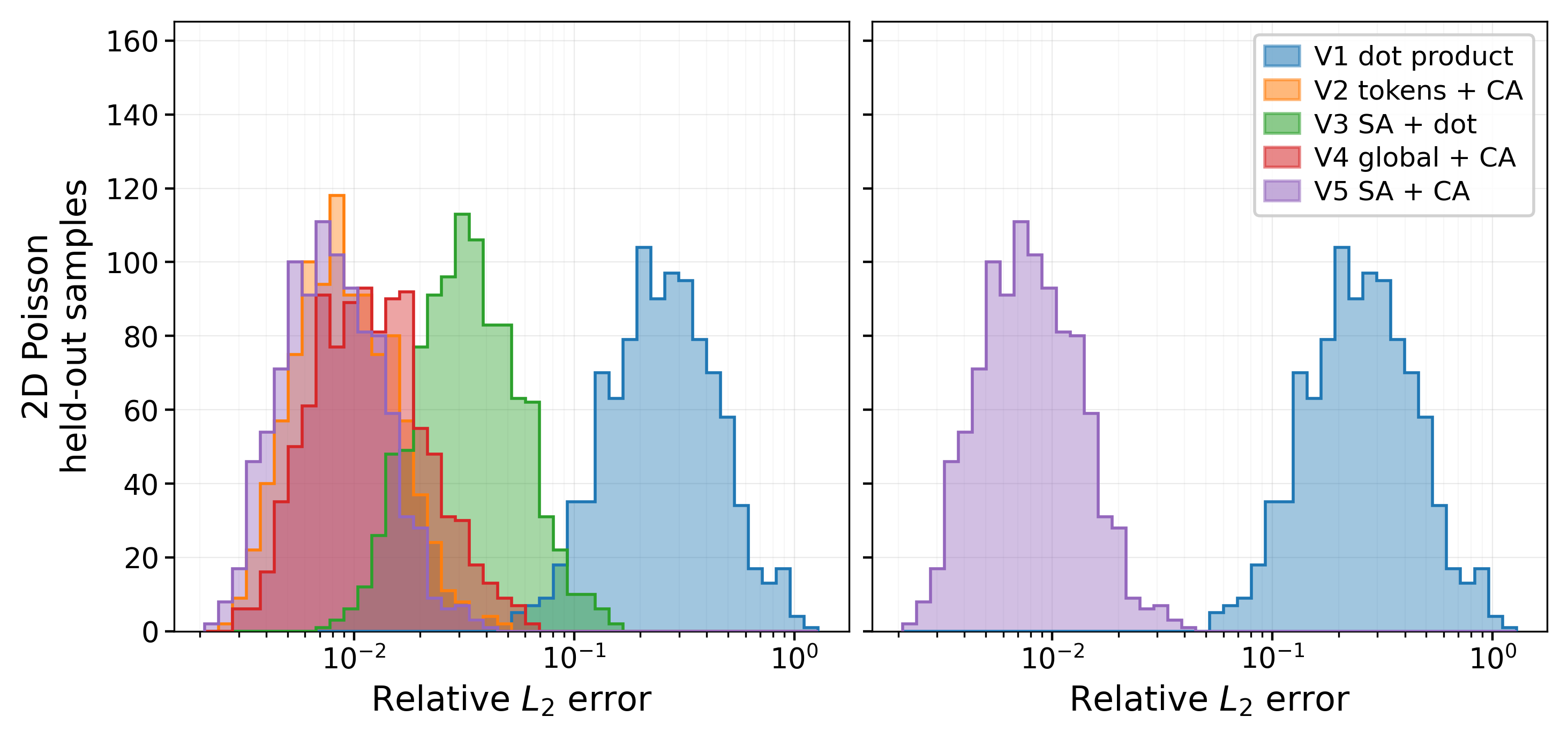}
\caption{Held-out error distributions for the data-driven Poisson models over 1000
matched test samples. Left: all five variants. Right: the V1 dot-product
baseline, mean relative $L_2$ error $2.95\times10^{-1}$, against the best variant
V5 at $9.13\times10^{-3}$.}
\label{fig:app-errhist-poisson-dd}
\end{figure}

\begin{figure}[pos=h!]
\centering
\includegraphics[width=0.74\textwidth]{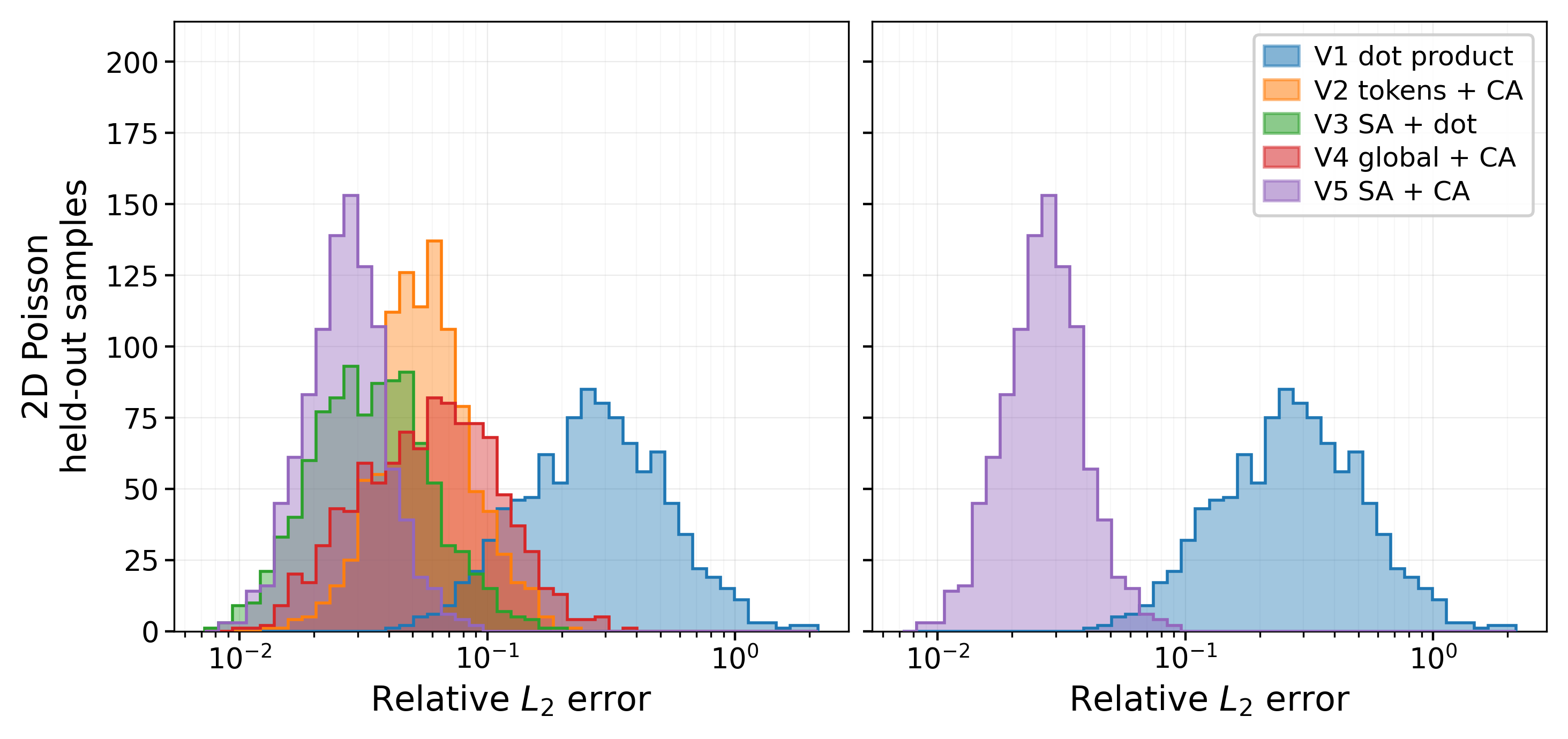}
\caption{Held-out error distributions for the physics-informed Poisson models over 1000
matched test samples. Left: all five variants. Right: the V1 dot-product
baseline, mean relative $L_2$ error $3.34\times10^{-1}$, against the best variant
V5 at $2.88\times10^{-2}$.}
\label{fig:app-errhist-poisson-pi}
\end{figure}

Figures~\ref{fig:app-errhist-dr-dd} to \ref{fig:app-errhist-poisson-pi} give the
distribution of the per-sample errors summarised by the means in
Table~\ref{tbl:main}.

\FloatBarrier

\printcredits

\bibliographystyle{cas-model2-names}
\bibliography{cas-refs}

\end{document}